\documentclass[10pt,twocolumn,letterpaper]{article}

\usepackage{cvpr}                

\usepackage{cuted}
\usepackage{amsmath}

\usepackage{graphicx}
\usepackage{booktabs}
\usepackage{pifont}
\usepackage{makecell}
\usepackage{adjustbox}
\usepackage{xcolor}

\newcommand{\cmark}{\textcolor{green!50!black}{\ding{51}}}
\newcommand{\xmark}{\textcolor{red!70!black}{\ding{55}}}

\definecolor{cvprblue}{rgb}{0.21,0.49,0.74}
\usepackage[pagebackref,breaklinks,colorlinks,allcolors=cvprblue]{hyperref}

\title{\emph{Insert In Style}: A Zero-Shot Generative Framework for Harmonious Cross-Domain Object Composition}

\author{Raghu Vamsi Chittersu, Yuvraj Singh Rathore, Pranav Adlinge, Kunal Swami \\
Samsung Research India Bangalore \\
Bengaluru, 560037, India \\
{\tt\small \{raghu.c, y.rathore, p.adlinge, kunal.swami\}@samsung.com}
}

\begin{document}
\maketitle
\addtocontents{toc}{\protect\setcounter{tocdepth}{-1}}

\begin{strip}
	\begin{minipage}{\textwidth}
		\centering
		\includegraphics[width=0.99\textwidth]{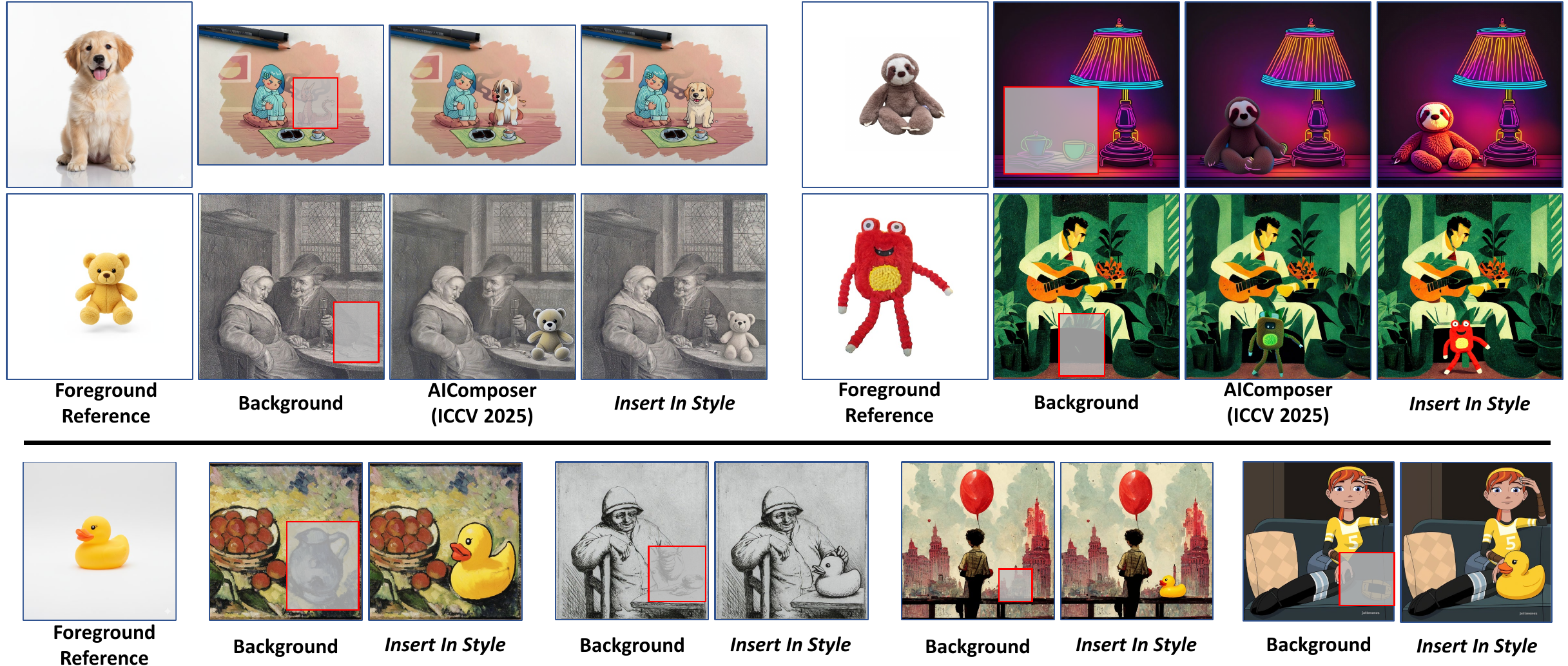}
		\captionof{figure}{\textbf{\emph{Insert In Style}: Zero-Shot Cross-Domain Composition.} (Rows 1-2) Comparison with the state-of-the-art cross-domain method AIComposer \cite{aicomposer_iccv2025}. AIComposer's ``blend-then-refine'' approach corrupts object identity by misapplying background features. \emph{Insert In Style} consistently generates a high-fidelity subject that is perfectly harmonized with the scene style. (Row 3) We demonstrate \emph{Insert In Style}'s versatile generalization: our single, zero-shot model seamlessly inserts one subject into four distinct stylized backgrounds.}
		\label{fig:teaser}
	\end{minipage}
\end{strip}

\begin{abstract}
Reference-based object composition involves integrating foreground reference image with background scene to produce harmonious fused image. This task becomes particularly challenging in cross-domain scenarios, where models must balance preserving the reference object's identity while harmonizing them to match stylized environments. This under-explored problem is currently split between practical ``blenders'' that lack generative fidelity and ``generators'' that require impractical, per-subject online finetuning. In this work, we introduce \textbf{Insert In Style}, the first zero-shot generative framework that is both practical and high-fidelity. Our core contribution is a unified framework with two key innovations: (i) a novel multi-stage training protocol that disentangles representations for identity, style, and composition, and (ii) a specialized masked-attention architecture that surgically enforces this disentanglement during generation (iii) A prior preservation objective that keeps learned identity and style priors intact. By design, this approach mitigates concept interference typical in unified-attention architectures while ensuring robust generalization across diverse references and styles. Our framework is trained on a new $115$k sample dataset, curated from a novel data pipeline. This pipeline couples large-scale generation with a rigorous, iterative human-in-the-loop filtering process to ensure both high-fidelity semantic identity and style coherence. Unlike prior work, our model is truly zero-shot and requires no text prompts. We also introduce a new public benchmark for stylized composition. We demonstrate state-of-the-art performance, significantly outperforming existing methods on both identity and style metrics, a result strongly corroborated by user studies.
\end{abstract}
    
\section{Introduction}
\label{sec:intro}

Reference-based object composition, focused on the task of inserting a specific object, identified by the reference image, in a scene at a specific location, identified by provided user mask,  \cite{dovenet_cvpr2020,objectplacement_neurips2018,deepharmonization_cvpr2017}, is a fundamental challenge in computer vision. Recent methods like DreamFuse \cite{dreamfuse_iccv2025}, AnyDoor \cite{anydoor_cvpr2024} and IMPRINT \cite{imprint_cvpr2024} have achieved remarkable realism. However, these models are trained almost exclusively on photorealistic data and fail spectacularly when composing objects into stylized domains like paintings, sketches, or digital art\textemdash a vast and common use case.

This cross-domain challenge has recently been met by two distinct families of methods. The first, \emph{``training-free blenders''}, includes pioneers like TF-ICON \cite{tficon_iccv2023} and, more recently, AIComposer \cite{aicomposer_iccv2025}. AIComposer \cite{aicomposer_iccv2025} represents the state-of-the-art for this class, cleverly removing the need for the precise text prompts that TF-ICON \cite{tficon_iccv2023} requires. These methods are fast and practical, but they are fundamentally blenders, not generators. They excel at harmonizing a pasted object but cannot generate a new object natively within the scene, limiting realism.

The second family, \emph{``online generators''} is represented by Magic Insert \cite{magicinsert_iccv2025}. This method achieves high generative fidelity by first finetuning a custom DreamBooth \cite{dreambooth_cvpr2023} model for a specific object, then performing style injection \cite{styleinjection_cvpr2024}. However, this quality comes at a prohibitive practical cost. Magic Insert \cite{magicinsert_iccv2025} is not zero-shot and requires a slow, computationally expensive, per-subject online finetuning process for every new object. This approach is impractical for real-world, drag-and-drop applications.

Concurrently, general-purpose controllers for DiT models, like OminiControl \cite{ominicontrol_iccv2025}, have proposed unified-attention mechanisms for handling multiple conditions. However, their effectiveness on complex, competing conditions \textemdash such as preserving identity while simultaneously transforming style \textemdash remains unproven.

The field is thus left with a clear gap: \emph{a method that is generative, zero-shot, and architecturally specialized for this competing-condition task}.

In this work, we introduce \emph{\textbf{Insert In Style}}, the \emph{first framework} to solve this challenge. Our core methodological contribution is two-fold: \emph{First}, we propose a multi-stage training protocol to explicitly disentangle representations: (a) a reference object encoder to learn robust identity, (b) a spatial style encoder to learn generalizable style, and (c) a final composition stage.  \emph{Second}, A novel approach to compose the trained competing conditions with attention disentanglement and prior preservation losses. This specialized architecture stands in direct contrast to general-purpose, unified-attention models and is key to balancing our competing objectives.

\begin{figure}[t!]
	\centering
	\includegraphics[width=0.99\linewidth]{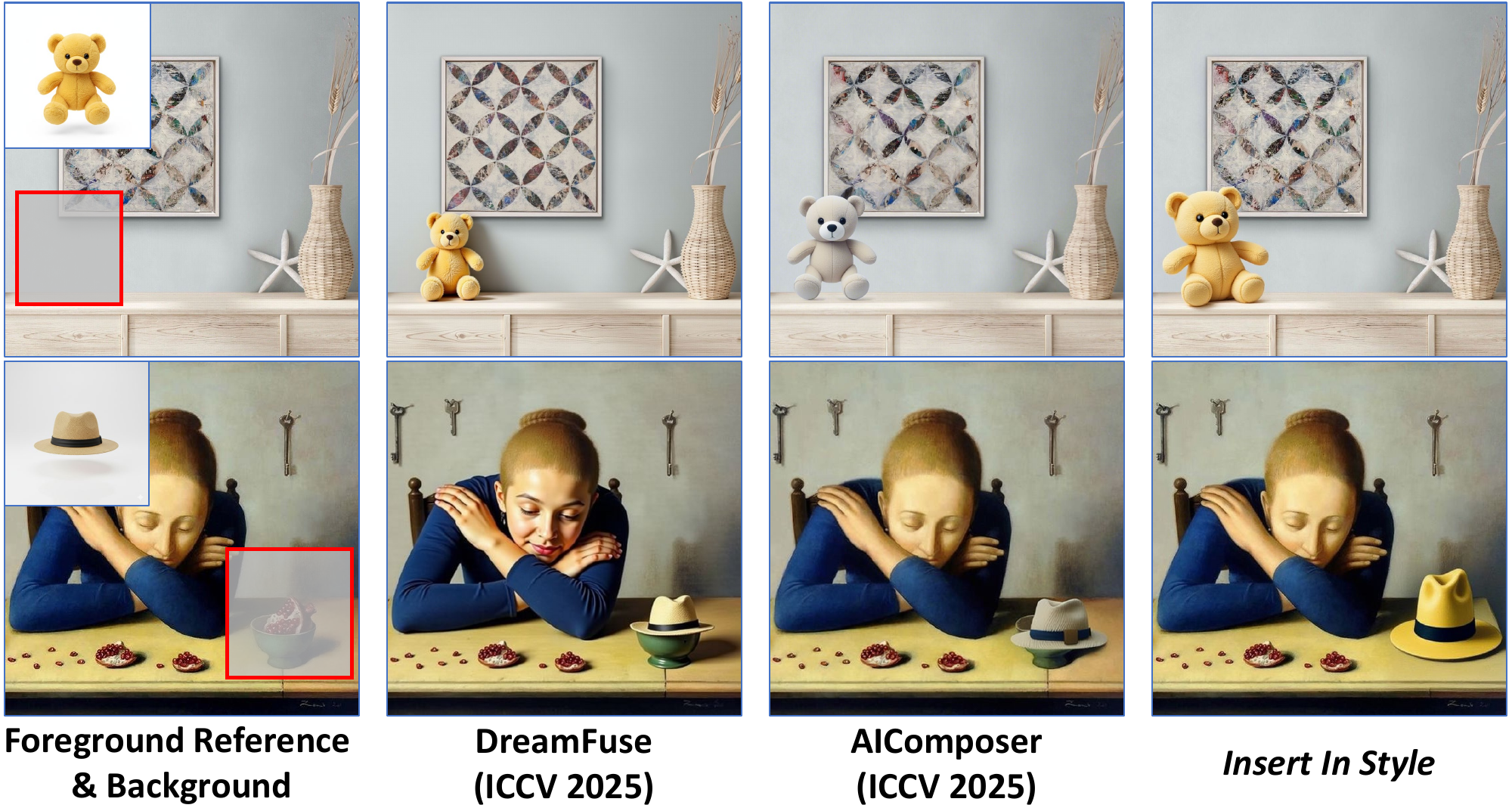}
	\caption{\emph{Insert In Style} generalizes across in-domain and cross-domain tasks. \textbf{Top (In-domain):} The cross-domain specialist method AIComposer \cite{aicomposer_iccv2025} incorrectly harmonizes the object. Our method maintains high fidelity, competitive with the in-domain specialist method DreamFuse \cite{dreamfuse_iccv2025}. \textbf{Bottom (Cross-domain):} DreamFuse \cite{dreamfuse_iccv2025} fails with a style mismatch, while AIComposer's \cite{aicomposer_iccv2025} harmonization corrupts object fidelity by incorrectly applying background style attributes. \emph{Insert In Style} uniquely generates a high-fidelity, style-coherent result.} 
	\label{fig:intro_motivation}
\end{figure}

To power this framework, we introduce a $115$k sample training corpus, created via a novel data pipeline that couples large-scale, multi-method generation with a robust, filtering process. Our iterative human-in-the-loop data filtering process involves ensemble of identity and style evaluation functions, which ensures our final dataset meets a high standard for both identity preservation and style coherence.

Our method is fully zero-shot at inference time. We demonstrate state-of-the-art performance on multiple cross-domain benchmarks, including our new \emph{Insert In Style Bench}, the largest and most comprehensive public benchmark we introduce for this task. Our method achieves a superior balance of identity preservation and style harmonization, a finding confirmed by extensive evaluation and user studies. Crucially, our model remains competitive on in-domain, photorealistic benchmarks, proving our framework extends a model's capabilities (see Fig.~\ref{fig:intro_motivation}).

Following are the major contributions of this work:
\begin{enumerate}
\item A novel generative framework featuring: (i) a three-stage training protocol that learns disentangled encoders for identity and style, and (ii) a masked-attention architecture that prevents feature-bleed between these competing conditions during composition (iii) a prior preservation objective that keeps learned identity and style priors intact.
\item The largest-scale dataset for this task ($115$k samples), curated by a novel, two-stage pipeline that is rigorously calibrated on human annotations to ensure both semantic identity and style coherence. We will make both the dataset and protocol public.
\item A new, diverse, and largest-scale public benchmark, \emph{Insert In Style Bench}, for evaluating cross-domain object composition, comprising $788$ samples spanning $51$ diverse background styles and $25$ subject categories.
\item State-of-the-art performance, outperforming baselines in quantitative, qualitative, and human evaluations. 
\end{enumerate}

\section{Related Work}
\label{sec:relatedwork}

\subsection{Generative Object Composition}
\label{subsec:generativeobjectcomposition}

\paragraph{In-domain Composition.}
In-domain composition focuses on realistically inserting an object into a photorealistic scene. Recent methods have excelled at preserving object identity. AnyDoor \cite{anydoor_cvpr2024} and MimicBrush \cite{mimicbrush_neurips2024} use specialized feature extractors, while IMPRINT \cite{imprint_cvpr2024} learns a dedicated identity-preserving representation. Other works leverage DiT \cite{dit_iccv2023} architectures, such as DreamFuse \cite{dreamfuse_iccv2025} and InsertAnything \cite{insertanything_arxiv2025}, for in-context editing, while ControlCom \cite{controlcom_arxiv2023} adds compositional control. While these methods achieve high fidelity in-domain, they are trained almost exclusively on photorealistic data and thus fail to generalize to stylized domains, creating jarring visual mismatches.

\paragraph{Cross-domain Object Composition.}
The challenge of cross-domain composition was first addressed by training-free ``blender'' methods. Pioneers like TF-ICON \cite{tficon_iccv2023} and its follow-ups, TALE \cite{tale_acmmm2024} and PrimeComposer \cite{primecomposer_acmmm2024}, manipulate diffusion latents and attention maps to harmonize a pasted object. The state-of-the-art in this class is AIComposer \cite{aicomposer_iccv2025}, which removes the reliance on precise text prompts. While these methods are fast and practical, they are fundamentally ``blend-then-refine'' approaches, not true generative models, limiting their realism.

A second family, ``online generators'', achieves higher fidelity. Magic Insert \cite{magicinsert_iccv2025} represents the state-of-the-art for this approach. It produces high-quality results by finetuning a custom DreamBooth \cite{dreambooth_cvpr2023} model per-subject. This quality, however, comes at a prohibitive practical cost: Magic Insert \cite{magicinsert_iccv2025} is not zero-shot and requires a slow, expensive, online finetuning process for every new object.

Thus, the field faces a clear trade-off: practicality (via AIComposer) versus generative fidelity (via Magic Insert). A framework that is both generative and zero-shot remains a critical open challenge that our work addresses.

\subsection{Controllable Diffusion Transformers (DiTs)}
\label{subsec:dits}
The advent of Diffusion Transformers (DiT) \cite{dit_iccv2023}  marked a shift from traditional UNets, with models like Stable Diffusion $3$ \cite{sd3_icml2024} and FLUX.$1$-dev \cite{fluxdev} establishing state-of-the-art performance. This created a need for parameter-efficient adaptation, solved by methods like LoRA \cite{lora_iclr2022}. OminiControl \cite{ominicontrol_iccv2025} emerged as the state-of-the-art general-purpose controller for DiTs, using a ``unified attention'' to process all conditions jointly.

\subsection{Style Transfer}
\label{subsec:styletransfer}
Style transfer is a well-studied field, with methods evolving from early neural approaches to modern \cite{omniconsistency_neurips2025,stylemaster_cvpr2025,styleinjection_cvpr2024,styleshot_arxiv2024,aespanet_iccv2023,cast_siggraph2022,artflow_cvpr2021}, high-fidelity diffusion-based techniques. Recent works like CSGO \cite{csgo_neurips2025} and OmniStyle \cite{omnistyle_cvpr2025} highlight the critical role of large-scale, high-quality data. However, these methods and their datasets are designed for style transfer, not object insertion. They lack the aligned foreground-reference and object-mask pairs essential for our task. This data gap has been the primary bottleneck for cross-domain composition. Our work is the first to address this by introducing \emph{Insert In Style}, a large-scale dataset with the precise \{\emph{foreground reference}, \emph{stylized scene}, \emph{foreground mask}\} triplets required.

\begin{figure*}[t!]
	\centering
	\includegraphics[width=0.99\linewidth]{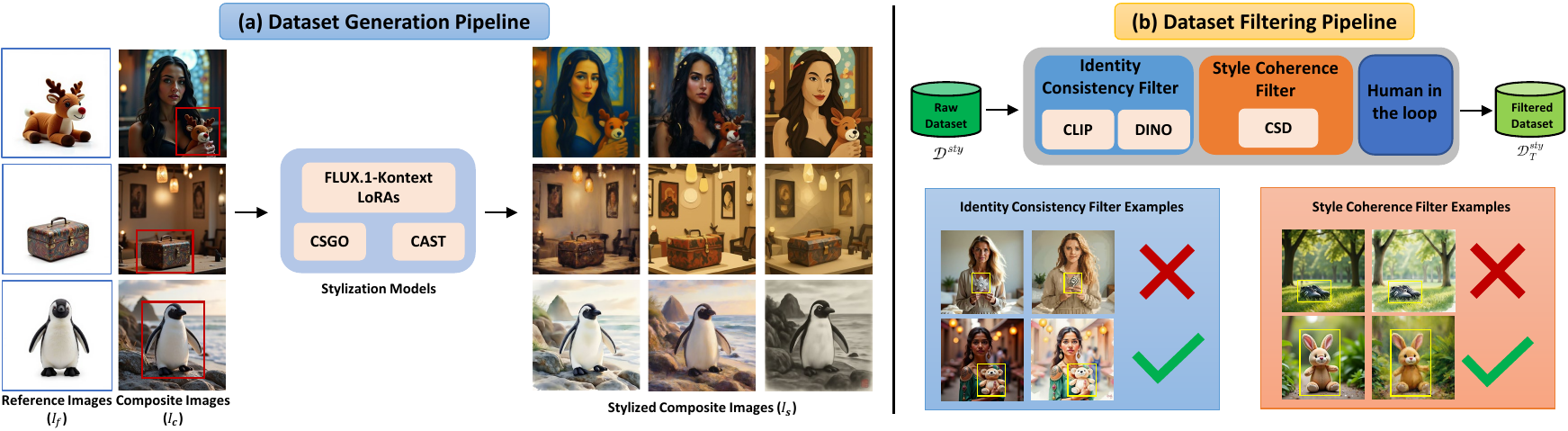}
	\caption{\textbf{Dataset Pipeline.} \textbf{(a) Generation:} We create a large-scale, diverse raw corpus by applying a mix of state-of-the-art stylization methods (FLUX.1-Kontext \cite{fluxkontextdev}, CSGO \cite{csgo_neurips2025}, and CAST \cite{cast_siggraph2022}). \textbf{(b) Filtering:} Our raw dataset is then refined by our rigorous iterative filtering process involving identity consistency evaluation and style coherence evaluation with human-in-the-loop (see Sec.~\ref{subsec:filteringmethods}).} 
	\label{fig:datasetpipeline}
\end{figure*}

\section{Dataset Generation}
\label{sec:datasetgeneration}
Traditional dataset curation approaches for reference-based object composition are limited to in-domain composition \cite{anydoor_cvpr2024, dreamfuse_iccv2025, mimicbrush_neurips2024}. To enable cross-domain compositional generalization, we propose a novel scalable approach built upon existing in-domain composition data. This section details our data curation methodology, which consists of a large-scale corpus generation and our novel filtering pipeline. The entire process is illustrated in Fig.~\ref{fig:datasetpipeline}. This data-centric approach is the foundation for our model's zero-shot generative capabilities.

\subsection{Data Generation Pipeline}
\label{subsec:datagenerationpipeline}
\paragraph{Base Data.}
Each sample in our dataset $\mathcal{D}$ originates from a triplet $\{I_f, I_c, I_m\}$, which includes:
\begin{itemize}
    \item a \textbf{foreground reference} image \( I_f \), serving as the object to be inserted.
    \item a \textbf{composite image} \( I_c \), representing the complete scene with the foreground object already inserted.
    \item a corresponding \textbf{binary mask} \( I_m \), indicating the region of the object in \( I_c \).
\end{itemize}

We build upon the DreamFuse dataset \cite{dreamfuse_iccv2025}, which curated diverse $\{I_f, I_c, I_m\}$ triplets through reference based generation. Upon inspection, we observed that a subset of these triplets contains a semantic mismatch between the reference $I_f$ and the object in the composite image $I_c$. To ensure the quality of our base data, we first proactively filter out these mismatched samples using CLIP-based similarity \cite{clipscore_emnlp2021} between the foreground reference $I_f$ and the masked object region in $I_c$. From this cleaned set, we then select foreground references from relevant classes (e.g., object, handheld, animal, pet, and product). This curation process yields our final base dataset, $\mathcal{D}^{base}$, of approximately $40,000$ high-quality triplets. Our core strategy for extending in-domain data to cross-domain involves performing diverse identity-preserving stylisation on in-domain composite images $I_c$ to generate its stylised variant $I_s$.

\paragraph{Generation of Stylized Variants.}
Our primary generative pipeline is built on FLUX.1-Kontext \cite{fluxkontextdev}, which we chose for its state-of-the-art performance in structure-preserving stylization \cite{fluxkontext_arxiv2025}, \cite{replicateblogfluxkontext}. We trained $20$ custom LoRA \cite{lora_iclr2022} modules for it on high-quality external style datasets \cite{sdxlstylesref} to create variants for popular and complex archetypes, such as \textit{Ghibli}, \textit{Watercolor}, and \textit{Chinese Ink}.

To broaden the style diversity beyond these $20$ archetypes and ensure our model generalizes, we supplemented this pipeline with state-of-the-art reference-based methods. We employed CSGO \cite{csgo_neurips2025} and CAST \cite{cast_siggraph2022}, which our empirical analysis showed are highly effective at preserving the subject's structural integrity and visual identity. We paired these methods with the \textit{Style$30$k} collection \cite{style30k_eccv2024}, a diverse benchmark covering $1,120$ fine-grained style categories. 

This combined generation process yields an initial corpus $\mathcal{D}^{sty}$ of approximately $350$k stylized compositions. Each sample in this final dataset consists of a $\{I_f, I_c, I_m, I_s\}$ quadruplet, where $I_s$ is the stylized composite image.

\subsection{Filtering Methods}
\label{subsec:filteringmethods}
This large, raw dataset inevitably contains a spectrum of failure cases, including, subject's semantic identity drift, and local style incoherence. A rigorous filtering stage is therefore essential to curate a high-quality corpus. We propose a novel, hybrid filtering pipeline to ensure both \emph{identity consistency} and \emph{style coherence}.

\paragraph{Identity Consistency Evaluation.}
After obtaining the raw stylized dataset $\mathcal{D}^{sty}$, it is important to ensure that the identity of the subject remains consistent between the composite image $I_c$ and the stylized composite image $I_s$. For this, we utilize the CLIP~\cite{clipscore_emnlp2021} score, which measures semantic similarity, and DINO~\cite{dinosv2_tmlr2024}, which measures structural similarity. First, we define an operation $\mathcal{C}(I, I_m)$ which, for a given image $I$ and mask $I_m$, crops the region of $I$ indicated by $I_m$. For every sample of $\mathcal{D}^{sty}$, we calculate $\phi_{\text{clip}}$ and $\phi_{\text{dino}}$ as follows:

\begin{align}
\phi_{\text{clip}} &= \text{CLIPSim}(\mathcal{C}(I_s, I_m), \mathcal{C}(I_c, I_m)) \\
\phi_{\text{dino}} &= \text{DINO}(\mathcal{C}(I_s, I_m), \mathcal{C}(I_c, I_m))
\end{align}

\begin{table*}[t!]
\centering
\caption{\textbf{Comparison with existing style and composition datasets, highlighting the data gap for cross-domain object composition.} Our \emph{Insert In Style Dataset} is the first large-scale corpus to provide all three components essential for this task: a foreground reference ($I_f$), a stylized composite image ($I_c^s$), and a ground-truth object mask ($I_m$). Style-focused datasets \cite{omnistyle_cvpr2025} lack object masks, while insertion-focused datasets \cite{dreamfuse_iccv2025} lack stylized scenes. Note that ``Ref.'' is short for ``Reference''.}
\scriptsize
\begin{adjustbox}{width=\textwidth}
\begin{tabular}{lccccccccc}
\toprule
\textbf{Dataset} & \textbf{Venue} & \textbf{Task Type} & \textbf{Foreground Ref.} & \textbf{Composite Image} & \textbf{Foreground Ref. Mask} & \textbf{Stylized Composite Image} & \textbf{\# Styles} & \textbf{\# Samples} \\
\midrule
Style-30K \cite{style30k_eccv2024}                      & ECCV $2024$       & Stylization                                & \xmark                    & \xmark & \xmark & \cmark         & $1120$                  & $30$k \\
Wiki-Art \cite{wikiart_arxiv2015}                       & arXiv $2015$      & Stylization                                & \xmark                    & \xmark & \xmark & \cmark         & $27$                    & $57$k \\
ArtBench \cite{artbench_arxiv2022}                      & arXiv $2022$      & Stylization                                & \xmark                    & \xmark & \xmark & \cmark         & $10$                    & $60$k \\
OmniConsistency \cite{omniconsistency_neurips2025}      & NeurIPS $2025$    & Stylization                                & \xmark                    & \cmark & \xmark & \cmark         & $22$                    & $2600$ \\
\midrule
OmniStyle \cite{omnistyle_cvpr2025}                     & CVPR $2025$       & Ref. based Stylization                     & \makecell{\xmark\\(has Style Ref.)}   & \cmark & \xmark & \cmark & $1000$              & $150$k\\
\midrule
DreamFuse \cite{dreamfuse_iccv2025}                     & ICCV $2025$       & Object Insertion                           & \cmark                    & \cmark & \cmark & \xmark         & -                       & $84$k \\
\midrule
\textbf{\emph{Insert In Style}}                         & -                 & \textbf{Cross-domain Object Insertion}     & \cmark                    & \cmark & \cmark & \cmark         & $\mathbf{1140}$         & $\mathbf{115}$\textbf{k} \\
\bottomrule
\end{tabular}
\end{adjustbox}
\label{tab:datasetcomparison}
\end{table*}

\paragraph{Style Coherence Evaluation}
While identity-based evaluation ensures semantic consistency, it does not guarantee stylistic coherence between the stylized object and its background. We observe that occasionally, stylization disproportionately affects some regions including the subject region, resulting in a perceptually inconsistent stylization between the subject and the rest of the image $I_s$. To ensure visual harmony, we use CSD \cite{csdmetric_eccv2024}, which measures the similarity of style characteristics between two images. We use CSD to calculate style similarity between the subject region and the remaining areas of the image $I_s$. For every sample $\{I_s, I_m\}$ from $\mathcal{D}^{sty}$, we calculate the CSD score:

\begin{equation}
\phi_{\text{csd}} = \text{CSD}( \mathcal{C}(I_s, I_m), \mathcal{P}(\mathcal{C}(I_s, 1-I_m))),
\end{equation}

where $\mathcal{P}$ is an operation that copies patches of size $64 \times 64$ from retained regions to masked-out regions. 

\paragraph{Iterative Human-in-the-loop Multidimensional Data Filtering}
To ensure the high alignment of our dataset across identity and style, we implement a \emph{progressive multidimensional filtering} strategy. We define an ensemble of scoring functions $\Phi = \{\phi_{\text{CLIP}}, \phi_{\text{DINO}}, \phi_{\text{CSD}}\}$ to evaluate the semantic, structural, and stylistic integrity of each instance $x \in \mathcal{D}^{sty}$.

The refinement process is formulated as an iterative distillation. At each iteration $t$, we identify a rejection set $\mathcal{R}_t$ comprised of the union of the lowest deciles across all scoring dimensions:
\begin{equation}
\mathcal{R}_t = \bigcup_{\phi \in \Phi} \{x \in \mathcal{D}_t \mid \phi(x) \leq Q_{0.1}(\phi)\}
\end{equation}
where $Q_{0.1}(\phi)$ represents the $10^{th}$ percentile of the score distribution for metric $\phi$ within the current subset $\mathcal{D}_t^{sty}$. The dataset is subsequently filtered via $\mathcal{D}_{t+1}^{sty} = \mathcal{D}_t^{sty} \setminus \mathcal{R}_t$.

Rather than employing a fixed iteration count, we adopt a human-centric convergence heuristic to balance dataset purity with sample diversity. After each pruning step, the rejection set $\mathcal{R}_t$ undergoes expert qualitative assessment. The process terminates at iteration $T$ when the samples in $\mathcal{R}_T$ exhibit high perceptual quality, specifically regarding identity preservation and stylistic fidelity. This filtering process reduces generated samples to the final $115$k high-quality, identity-preserving, and style-coherent training corpus. We provide more details on the dataset generation, filtering process and training data composition in the \emph{supplementary}. Tab.~\ref{tab:datasetcomparison} shows a detailed comparison to existing datasets establishing our corpus as the first and largest to provide the aligned $\{I_f, I_c, I_m, I_s\}$ quadruplets essential for this task.

\begin{figure*}[t!]
	\centering
	\includegraphics[width=0.99\linewidth]{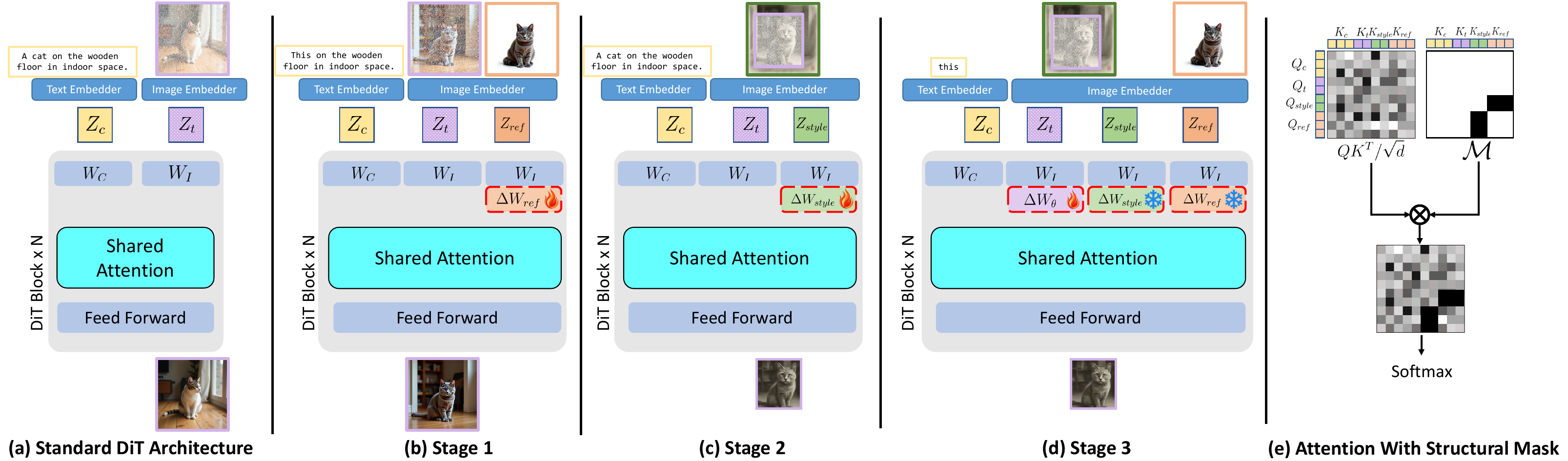}
	\caption{\textbf{Our multi-stage training protocol on a DiT backbone (a).} Stages $1$ (b) and $2$ (c) are trained in parallel to independently learn object and style encoding. Stage-$3$ (d) learns composition by assembling these frozen branches, guided by our Structural Mask Attention (e).} 
	\label{fig:modelarchitecture}
\end{figure*}

\section{Proposed Method}
\label{sec:proposedmethod}
The core challenge in cross-domain object composition is to balance two competing objectives: identity preservation (which requires preserving the object's features) and style harmonization (which requires \textit{transforming} them). This presents a unique challenge for general purpose multi condition adapter fusion methods \cite{ominicontrol_iccv2025}, where all independently trained conditions are trained with same generative objective. This joint processing, while powerful, is not explicitly designed to manage competing signals, creating a potential risk of \emph{concept interference} or \emph{feature bleed}, where the strong style signal may corrupt identity features, or vice-versa. To solve this, we introduce a novel framework that is a two-fold contribution:
\begin{enumerate}
    \item A multi stage training protocol that explicitly disentangles these competing concepts by pre-training specialized, independent encoders for identity and style objectives.
    \item A novel approach to compose the pre-trained competing conditions with attention disentanglement and prior preservation.
\end{enumerate}

\subsection{Model Architecture}
\label{subsec:modelarchitecture}
\paragraph{Base Model.} Our proposed framework built atop pre-trained DiTs, using FLUX.$1$-dev as our primary base model \cite{fluxdev}. Notably, our formulation is backbone agnostic and can be extended to modern T2I or instruction-based editing DiTs \cite{fluxkontext_arxiv2025,dit_iccv2023}. FLUX.$1$-dev operates on latent representations: a VAE \cite{vae_iclr2014} encodes images into latents $Z_0 \in \mathbb{R}^{H \times W \times C}$, and a T$5$ encoder \cite{t5transformer_jmlr2020} produces text embeddings $Z_c \in \mathbb{R}^{L \times D}$. The model is trained as a rectified flow \cite{flowmatching_iclr2023} to predict the velocity vector $v_\theta$ of a flow matching a linear interpolation between noise $Z_1 \sim \mathcal{N}(0, I)$ and the target image latent $Z_0$. We define the path as $Z_t = t \cdot Z_0 + (1-t) \cdot Z_1$, where $t \in [0, 1]$. The target velocity is $v^* = Z_0 - Z_1$, and the flow matching objective:

\begin{gather}
\mathcal{L}_{\text{flow}} =
\mathbb{E}_{t,\,Z_0,\,Z_1}\left[
\left\|
v_\theta(Z_t, c, t)
-
v^*
\right\|_2^2
\right]
\label{eq:flowmatchingloss}
\end{gather}

\paragraph{Disentangled Conditioning Architecture.}
To handle our competing conditions, we extend the FLUX.$1$-dev \cite{fluxdev} architecture with two additional, parameter-efficient conditioning branches {\cite{ominicontrol_iccv2025} . The full model processes four parallel token sequences:

\begin{enumerate}
    \item \textbf{Image Latents ($Z_t$):} The noisy image tokens to be denoised.
    \item \textbf{Text Embeddings ($Z_c$):} Standard text prompt conditioning.
    \item \textbf{Identity Branch ($Z_{ref}$):} A new branch to encode the foreground reference object.
    \item \textbf{Style Branch ($Z_{style}$):} A new branch to encode the background style and spatial context.
\end{enumerate}

We initialize the \textbf{Identity} and \textbf{Style} branches with the same architecture and weights as the base FLUX.$1$-dev \cite{fluxdev} image branch. We then insert LoRA \cite{lora_iclr2022} adapters into all QKV projections and MLP layers of these two new branches, as well as the main image branch. Only these LoRA parameters are trained.

In each Transformer block, all four token sequences are jointly processed. The QKV matrices for the new conditional branches are computed using the shared weights $W_I^{Q,K,V}$ plus their branch-specific LoRA adapters, $\Delta W$:

\begin{gather}
\begin{aligned}
    &Q_{\text{ref}} = Z_{\text{ref}}^h (W_I^Q + \Delta W^Q_{\text{ref}}), \\
    &K_{\text{ref}} = Z_{\text{ref}}^h (W_I^K + \Delta W^K_{\text{ref}}), \\
    &V_{\text{ref}} = Z_{\text{ref}}^h (W_I^V + \Delta W^V_{\text{ref}})
\end{aligned} \\
\begin{aligned}
    &Q_{\text{style}} = Z_{\text{style}}^h (W_I^Q + \Delta W^Q_{\text{style}}), \\
    &K_{\text{style}} = Z_{\text{style}}^h (W_I^K + \Delta W^K_{\text{style}}), \\
    &V_{\text{style}} = Z_{\text{style}}^h (W_I^V + \Delta W^V_{\text{style}})
\end{aligned}
\end{gather}

These are then concatenated with the image ($Q_t, K_t, V_t$) and text ($Q_c, K_c, V_c$) tokens for the shared self-attention operation \cite{ominicontrol_iccv2025}.

\subsection{Conditional Branch Training}
\label{sec:training}
We train the LoRA adapters of our architecture to incrementally build and align the required representations. Our protocol uses the same frozen, pre-trained FLUX.$1$-dev \cite{fluxdev} model weights as the foundation for three distinct training stages. We first train two specialist encoders independently and in parallel, then assemble them to train a final \emph{compose} model \emph{Insert In Style}.

\paragraph{Reference Object Encoder.}
We first attach a new \textbf{Identity Branch} ($Z_{ref}$) to the frozen, pre-trained FLUX.$1$-dev \cite{fluxdev} base model and train only its LoRA adapters 
($\Delta W_{\text{ref}}$). The model is trained on the Subjects$200$K Collection $2$ dataset \cite{ominicontrol_iccv2025}, which contains paired reference objects and their corresponding composed scenes. We use text prompts that implicitly reference the subject (e.g., ``a photo of \textit{this item} on a table''), forcing the model to ground the subject's identity in the visual features of $Z_{ref}$. The model is trained to reconstruct the full scene using $\mathcal{L}_{\text{flow}}$ (Eq.~\ref{eq:flowmatchingloss}), conditioned on $Z_t$, $Z_{ref}$, and $Z_c$.

\paragraph{Spatial Style Encoder.}
Independently, we attach a new \textbf{Style Branch} ($Z_{style}$) to the same frozen, pre-trained FLUX.$1$-dev \cite{fluxdev} base model and train only its LoRA adapters ($\Delta W_{\text{style}}$). This parallel process ensures the style representations are learned completely independently from the identity representations. The model is trained on a diverse corpus of $70,000$ images, including $40,000$ stylized scenes from OmniStyle \cite{omnistyle_cvpr2025}, $15,000$ from StyleBooth \cite{stylebooth_iccvw2025}, and $15,000$ real-world images. We train the model on a style-aware inpainting task. Given an image latent $Z_i$ and a binary mask $M$, we define the style context as the unmasked tokens $Z_{\text{style}} = Z_i \odot (1-M)$ and the noisy target tokens as $Z_t$ (the masked region $Z_i \odot M$, noised). The model is trained to denoise $Z_t$ conditioned on $Z_{\text{style}}$ and a text prompt $Z_c$.

\subsection{Insert In Style: Style Aware Reference Composition}
\label{subsec:iis_styleawarereferencecomposition}
To formulate the complete compositional framework, we integrate the pre-trained, Identity Branch ($\Delta W_{\text{ref}}$) and Style Branch ($\Delta W_{\text{style}}$) into the FLUX.1-dev \cite{fluxdev} base model. As these branches are optimized for disparate objectives, standard inference-time adapter mixing \cite{ominicontrol_iccv2025} yields sub-optimal performance. Consequently, we introduce a novel set of trainable LoRA adapters, $\Delta W_{\theta}$, applied to the main branch ($Z_t$). This compositional stage is trained using our proposed \emph{Insert In Style} dataset, comprising $115$k curated samples (see Sec.~\ref{sec:datasetgeneration}).

However, na\"ively fine-tuning the unified model inherently compromises the generalization capabilities of the individual constituent branches. This degradation arises fundamentally from the joint optimization process itself, forcing previously isolated semantic spaces to update simultaneously leads to severe feature entanglement, independent of the composite data distribution. To mitigate this structural interference, we propose Masked Attention to explicitly decouple the \textbf{Identity} and \textbf{Style} branches within the multi-modal attention mechanism. Furthermore, directly finetuning LoRA weights for complex compositional tasks inherently risks over-entangling the structural and stylistic conditions, leading to catastrophic forgetting of the base model's broad generative priors. Therefore, we incorporate a prior preservation term into our training objective (Eq.\ref{eq:prior_loss_term}).

\paragraph{Structural Attention Mask} To prevent the \emph{concept interference} between our two competing conditions, we introduce a \textbf{structural attention mask $\mathcal{M}$}. This mask is applied during the shared self-attention calculation to surgically control information flow, as shown in Fig.~\ref{fig:modelarchitecture}. We define the concatenated query, key, and value matrices as:

\begin{gather}
\begin{aligned}
    Q = [Q_c; Q_t; Q_{\text{style}}; Q_{\text{ref}}] \\
    K = [K_c; K_t; K_{\text{style}}; K_{\text{ref}}] \\
    V = [V_c; V_t; V_{\text{style}}; V_{\text{ref}}]
\end{aligned}
\end{gather}

The full attention operation then becomes:

\begin{gather}
S = \text{softmax} \left( \frac{Q K^\top}{\sqrt{d}} + \mathcal{M} \right) \\
[Z_c^{h+1}; Z_t^{h+1}; Z_{\text{style}}^{h+1}; Z_{\text{ref}}^{h+1}] = S V
\end{gather}

The mask $\mathcal{M}$ (a matrix of $0$s and $-\infty$) is configured to enforce two rules: (i) all branches can attend to the text ($Z_c$) and image ($Z_t$) tokens, but (ii) the \textbf{Identity Branch ($Z_{ref}$)} and \textbf{Style Branch ($Z_{style}$)} are masked from attending to each other. This enforces the disentanglement learned in Stages $1$ and $2$, allowing the model to compose the object harmoniously without the style signal \emph{bleeding} into and corrupting the identity signal, or vice-versa. We ablate this key design choice in Sec.~\ref{subsec:ablationstudy}.

\paragraph{Identity Prior Preservation.}
We enforce identity prior preservation through a controlled dropout mechanism. During final training, we randomly omit the style reference context with a fixed probability $p_{\text{drop}} = 0.2$, substituting it with a neutral, generic real-world image representation that contains no specific style context. This forces the model to maintain high fidelity to the identity reference, preventing it from relying purely on the style to dictate structure, which is a crucial property for authentic real-domain insertion.

\paragraph{Style Prior Preservation.}
To ensure that the style branch's core representational knowledge is maintained and to prevent catastrophic drift towards identity preservation at the expense of stylistic fidelity, we introduce a custom style prior loss, $\mathcal{L}_{\text{prior}}^{\text{style}}$. We formulate this regularizer directly in the velocity prediction space by minimizing the discrepancy between the output of the evolving composite model and the target output of the frozen, style-specific pre-trained model. Formally, this objective is defined as:

\begin{equation}
\mathcal{L}_{\text{prior}}^{\text{style}} = \mathbb{E}_{z_t, t, c, z_{\text{sty}}, z_{\text{ref}}} \left[ \left\| v_{\theta}(z_t, c, z_{\text{sty}}, z_{\text{ref}}) - v_{\text{style}}(z_t, c, z_{\text{sty}}) \right\|_2^2 \right]
\label{eq:prior_loss_term}
\end{equation}

where $v_{\theta}(z_t, c, z_{\text{sty}}, z_{\text{ref}})$ denotes the velocity predicted by our final compositional model parameterized by $\theta$, conditioned on the noisy latent $z_t$, text prompt $c$, style context $z_{\text{sty}}$, and identity reference $z_{\text{ref}}$. Correspondingly, $v_{\text{style}}(z_t, c, z_{\text{sty}})$ represents the reference velocity generated by the frozen style model, which observes only the style and text conditioning. This loss forces the model to harmonize identity insertion without overwriting its foundational style priors. We train the final composite model with the objective $\mathcal{L}_{\text{flow}} + \alpha \mathcal{L}^{\text{style}}_{\text{prior}}$, where $\alpha$ is set to $0.2$.

\begin{figure*}[t!]
	\centering
    \includegraphics[width=0.98\linewidth]{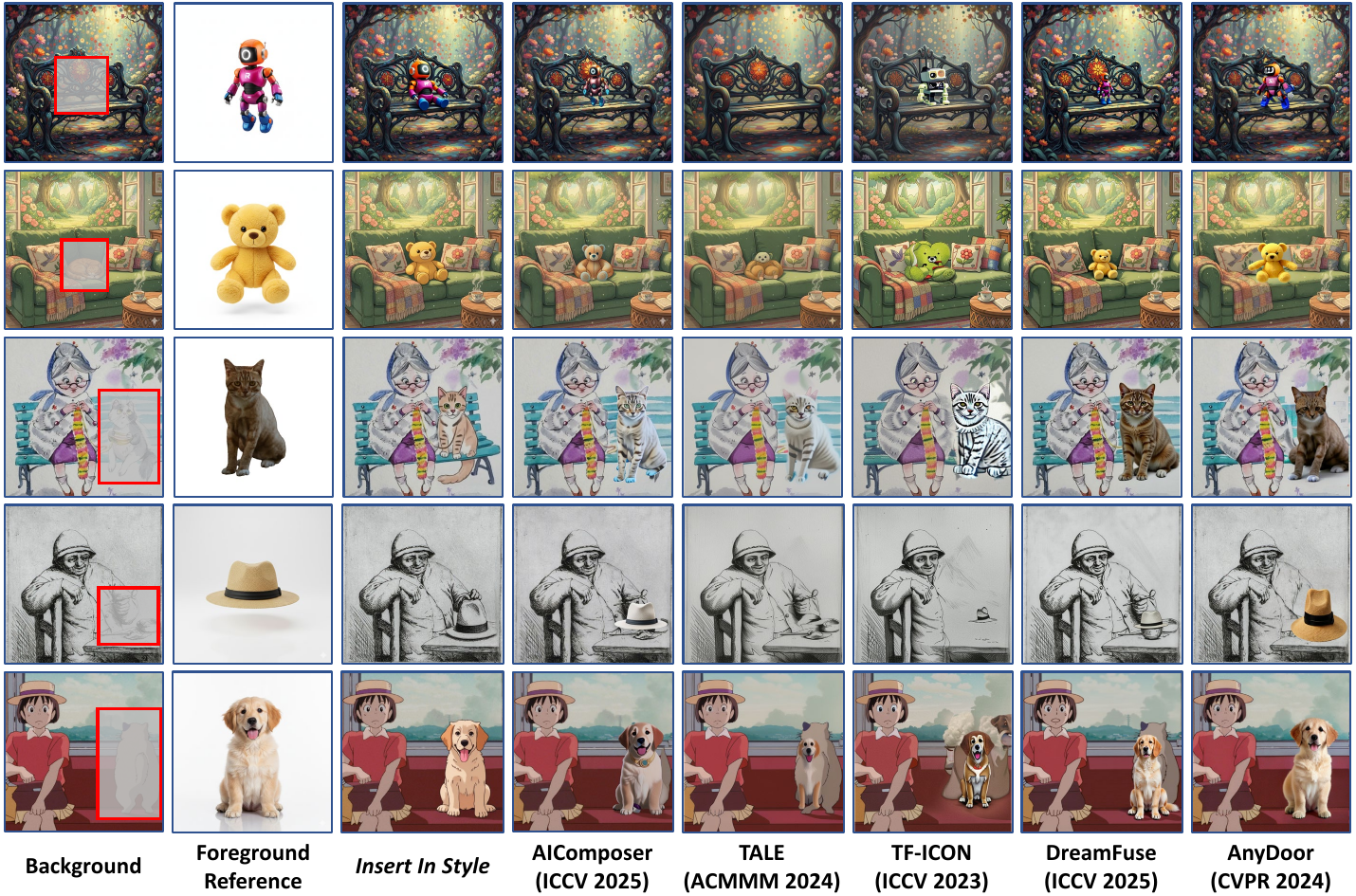}
	\caption{\textbf{Qualitative comparison with state-of-the-art in-domain and cross-domain baselines.} In-domain methods \cite{dreamfuse_iccv2025, anydoor_cvpr2024} produce jarring style mismatches, failing to generalize. Cross-domain methods \cite{aicomposer_iccv2025, tale_acmmm2024, tficon_iccv2023} corrupt the subject's identity and fidelity. In contrast, \emph{Insert In Style} consistently achieves a superior balance, producing results that are both high-fidelity and aesthetically harmonious.} 
	\label{fig:qualitativecomparison_main}
\end{figure*}

\section{Experiments}
\label{sec:experiments}

\subsection{Implementation Details}
\label{subsec:implementationdetails}
We utilize the PyTorch framework \cite{pytorch_neurips2019} and train all LoRA modules initialized with rank $16$ using Prodigy optimizer \cite{prodigy_icml2024} with  learning rate of $1.0$ for $1$ epoch. All our experiments are conducted on $4$ NVIDIA A$100$ GPUs, using a gradient accumulation factor of $2$, resulting in effective batch size of $8$. Both training and inference are performed at spatial resolution of $768\times768$ pixels.

\subsection{Evaluation Benchmarks} 
\label{subsec:evaluationbenchmarks}
\paragraph{AIComposer Benchmark}
We benchmark our method on the AIComposer dataset \cite{aicomposer_iccv2025}, which aggregates $367$ background-foreground pairs and incorporates the $95$ cross-domain samples from the TF-ICON benchmark \cite{tficon_iccv2023}. The benchmark provides a rigorous test of generalization, featuring a wide array of background styles (e.g., \textit{Sketch}, \textit{Watercolor}, \textit{Sci-Fi}, \textit{Pixel Art}) and diverse foreground categories (e.g., \textit{Animals}, \textit{Food}, \textit{Buildings}, \textit{Cartoon subjects}).

\paragraph{Insert In Style-Bench}
To rigorously evaluate generalization, we introduce the novel \emph{Insert In Style Bench}. To our knowledge, it is the \emph{largest evaluation benchmark for this task}, comprising $788$ challenging pairs. It is specifically designed to test the insertion of photorealistic objects into complex, stylized scenes. It pairs $25$ diverse foreground concepts (e.g., pets, food, toys), sourced from generative models \cite{gemini25_arxiv2025} and the Dreambooth dataset \cite{dreambooth_cvpr2023}, with $51$ highly varied backgrounds. To ensure broad stylistic diversity, the backgrounds are meticulously curated from public sources, including Human-Art \cite{humanart_cvpr2023}, the Wikiart dataset \cite{wikiart_arxiv2015}, Kaggle datasets \cite{kaggledb1,kaggledb2,kaggledb3,kaggledb4,kaggledb5}, and Pexels \cite{pexels}.

\begin{table}[t!]
\centering
\caption{Quantitative comparison on AIComposer benchmark dataset. The best results are in \textbf{bold}, and the second best are \underline{underlined}.}
\resizebox{0.48\textwidth}{!}{
    \begin{tabular}{lccccccc}
    \toprule
    \textbf{Method}                 & \textbf{Venue} & \textbf{CLIP-I $\uparrow$} & \textbf{CSD $\uparrow$} & \textbf{AES $\uparrow$} & \textbf{Overall Mean $\uparrow$} \\
    \midrule
    AnyDoor                         & CVPR $2024$    & \textbf{0.831}      & 0.382               & 0.611              & 0.608 \\
    DreamFuse                       & ICCV $2025$    & \underline{0.784}   & 0.458               & 0.632              & 0.625 \\ 
    \midrule
    TF-ICON                         & ICCV $2023$    & 0.714               & 0.438               & 0.584              & 0.579 \\
    TALE                            & ACMMM $2024$   & 0.686               & \textbf{0.495}      & 0.607              & 0.596 \\
    AIComposer                      & ICCV $2025$    & 0.774               & 0.476               & \underline{0.644}  & \underline{0.631} \\
    \midrule
    \emph{Insert In Style}          & -              & 0.779               & \underline{0.481}   & \textbf{0.655}     & \textbf{0.638} \\
    \bottomrule
    \end{tabular}
}
\label{tab:benchmark_aicomposer}
\end{table}

\begin{table}[t!]
\centering
\caption{Quantitative comparison on \emph{Insert In Style Bench} dataset. The best results are in \textbf{bold}, and the second best are \underline{underlined}.}
\resizebox{0.48\textwidth}{!}{
    \begin{tabular}{lccccccc}
    \toprule
    \textbf{Method}                 & \textbf{Venue} & \textbf{CLIP-I $\uparrow$} & \textbf{CSD $\uparrow$} & \textbf{AES $\uparrow$} & \textbf{Overall Mean $\uparrow$} \\
    \midrule
    AnyDoor                         & CVPR $2024$    & \textbf{0.863}      & 0.318               & 0.656              & 0.612 \\
    DreamFuse                       & ICCV $2025$    & 0.758               & 0.449               & 0.681              & 0.629 \\ 
    \midrule
    TF-ICON                         & ICCV $2023$    & 0.687               & 0.382               & 0.661              & 0.577 \\
    TALE                            & ACMMM $2024$   & 0.671               & \underline{0.462}     & \underline{0.695}    & 0.609 \\
    AIComposer                      & ICCV $2025$    & \underline{0.768}   & 0.430               & 0.692                & \underline{0.630} \\
    \midrule
    \emph{Insert In Style}          & -              & 0.761               & \textbf{0.466}        & \textbf{0.697}     & \textbf{0.641} \\
    \bottomrule
    \end{tabular}
}
\label{tab:benchmark_ours}
\end{table}

\subsection{Comparison with Existing Methods}
\label{subsec:comparisonwithsota}
We compare our method against state-of-the-art in-domain object insertion baselines (DreamFuse \cite{dreamfuse_iccv2025}, AnyDoor \cite{anydoor_cvpr2024}) and cross-domain composition methods (TF-ICON \cite{tficon_iccv2023}, TALE \cite{tale_acmmm2024}, AIComposer \cite{aicomposer_iccv2025}). We present comprehensive qualitative and quantitative results on the benchmarks detailed in Sec.~\ref{subsec:evaluationbenchmarks}.

We evaluate all methods across three key aspects: \emph{identity preservation}, \emph{style consistency}, and \emph{aesthetic quality}. We measure \emph{identity preservation} using CLIP-I \cite{clipscore_emnlp2021} between the reference image and the edited region. \emph{Style consistency} is quantified via CSD \cite{csdmetric_eccv2024} between the edited region and the background. \emph{Aesthetic quality} is measured using a pre-trained Aesthetic Score (\textsc{AES}) model \cite{laion5b_neurips2022}. Crucially, CSD and \textsc{AES} are calculated only when the edit mask (found via pixel differencing and threshold) exceeds $20\%$ of the image. This prevents a known bias where methods that fail to make an edit are unfairly rewarded. We emphasize that relying on any single metric may not fully capture a method's effectiveness; thus, we introduce an Overall Mean across metrics for comprehensive evaluation.

Quantitative results are presented in Tab.~\ref{tab:benchmark_aicomposer} and Tab.~\ref{tab:benchmark_ours} for AIComposer benchmark \cite{aicomposer_iccv2025} and our \emph{Insert In Style Bench}, respectively. The results reveal a clear and consistent trade-off in the state-of-the-art: in-domain models (DreamFuse \cite{dreamfuse_iccv2025}, AnyDoor \cite{anydoor_cvpr2024}) achieve high identity (CLIP-I) but fail on style (low CSD/AES). Conversely, cross-domain methods (AIComposer \cite{aicomposer_iccv2025}, TALE \cite{tale_acmmm2024}, TF-ICON \cite{tficon_iccv2023}) achieve better stylization but lose object identity (low CLIP-I). Our method successfully resolves this dilemma. We simultaneously achieve high identity preservation, strong style consistency, and superior aesthetic quality, outperforming all baselines on this challenging task.

Qualitative results are presented in Fig.~\ref{fig:qualitativecomparison_main}. These comparisons visually confirm the quantitative findings: in-domain methods like AnyDoor \cite{anydoor_cvpr2024} produce jarring stylistic mismatches. Cross-domain methods like AIComposer \cite{aicomposer_iccv2025} corrupt the subject's identity, causing visual artifacts and misapplying background features. Our method consistently achieves a superior balance, producing results that are both high-fidelity and aesthetically harmonious.

Additional qualitative and quantitative results on SubjectPlop benchmark (by Magic Insert \cite{magicinsert_iccv2025}) and in-domain object composition on AnyDoor \cite{anydoor_cvpr2024} benchmark are provided in the \emph{supplementary} along with comparisons against state-of-the-art instruction-based editing models.

\begin{figure}[t!]
	\centering
    \includegraphics[width=0.98\linewidth]{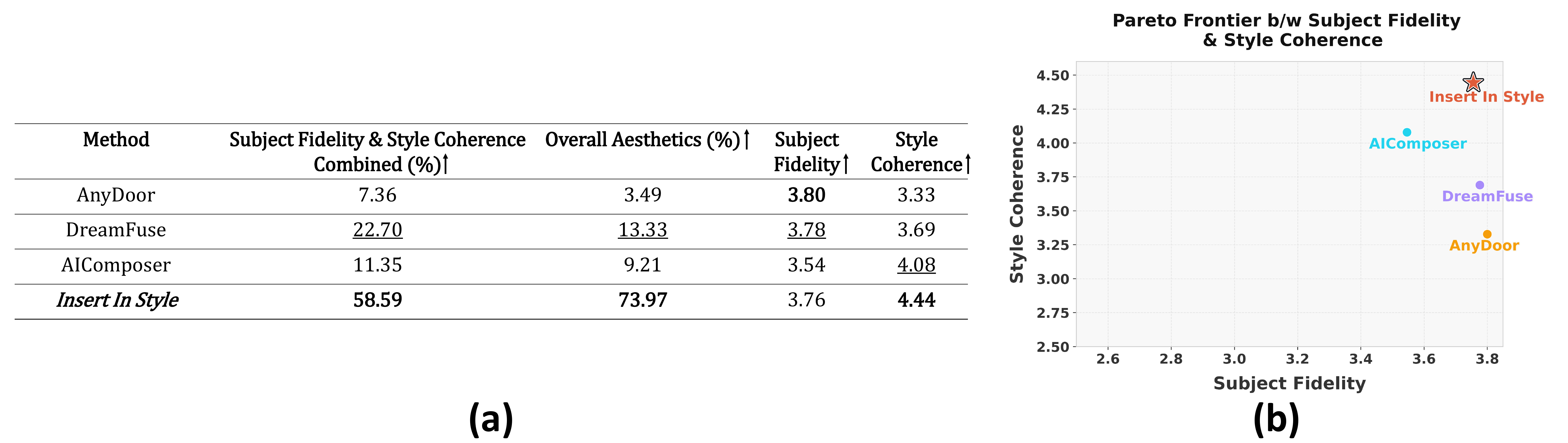}
	\caption{\textbf{User study.} In a randomized and blind comparative study, \emph{Insert In Style} was strongly preferred for ``Subject Fidelity, Style Coherence and overall aesthetic quality.''.} 
	\label{fig:userstudy}
\end{figure}

\subsection{User Study}
\label{subsec:userstudy}
To evaluate perceptual quality, a user study with $33$ participants was conducted. In \textbf{Fig.~\ref{fig:userstudy} (a)} Participants were asked to rank best output based on (i) combined subject fidelity and style coherence, (ii) overall aesthetic quality, results shown in first 2 columns. Users prefer our method based on these holistic evaluations over other baselines. Further, we requested users to score anonymized, randomized results on subject fidelity and style coherence separately ($0.0$-$5.0,5.0$ being best), shown in last 2 columns and also visualised through the \emph{pareto frontier} \textbf{Fig.~\ref{fig:userstudy} (b)} showcasing \emph{Insert In Style} handles the balance between identity and styles better than existing methods.

\subsection{Ablation Study}
\label{subsec:ablationstudy}
We validate our complete methodology in Tab.~\ref{tab:ablationstudy} and Fig.~\ref{fig:ablationstudy}. The \emph{Na\"ive E2E} variant (Row $1$) exhibits a catastrophic failure, inserting random objects (as seen in Fig.~\ref{fig:ablationstudy}) and achieving the lowest Overall Mean. This shows our multi-stage protocol is necessary. Tab.~\ref{tab:ablationstudy} reveals a clear Identity-Style trade-off. The \emph{w/o Style pre-train} variant (Row $3$) achieves the highest CLIP-I but suffers the worst CSD. Conversely, adding the style pre-train without our mask (Row $4$) improves CSD but hurts CLIP-I. This demonstrates the competing objectives and ``concept interference'' that we aim to solve. \emph{Insert In Style} (Row $5$), by adding the masked-attention, solves this trade-off: compared to the \emph{w/o Masked Attention} variant (Row $4$), \emph{Insert In Style} simultaneously improves CLIP-I, CSD, and AES, achieving the best Overall Mean score.

In Tab.~\ref{tab:ablationstudy_priorloss}, we demonstrate the effectiveness of our prior preservation objective for identity and style. The decline in CLIP-I performance \emph{w/o Identity prior preservation} v/s the baseline \emph{Insert In Style} confirms the identity prior's role in retaining reference object identity during stylization. Likewise, the reduced CSD for \emph{w/o Style Prior Preservation} highlights the style prior's contribution to harmonized results. Additional ablations considering data filtering and training strategies are  provided in the \emph{supplementary}.

\begin{figure}[t!]
	\centering
	\includegraphics[width=0.98\linewidth]{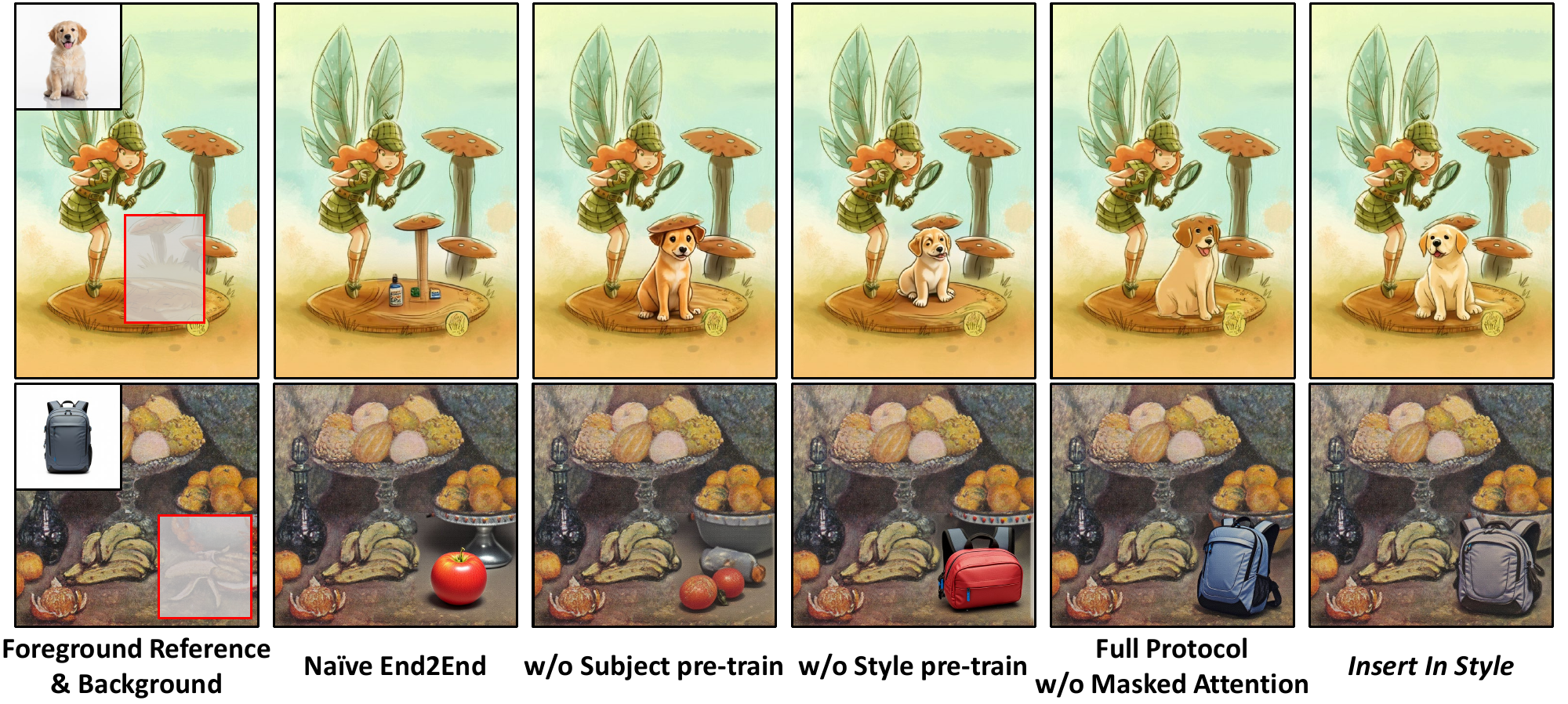}
	\caption{Qualitative results of the ablation study on our multi-stage training protocol and masked-attention architecture.} 
	\label{fig:ablationstudy}
\end{figure}

\begin{table}[t!]
\centering
\caption{Ablation on our multi-stage training protocol and masked-attention architecture.}
\resizebox{0.48\textwidth}{!}{
    \begin{tabular}{lcccccc}
    \toprule
    \textbf{Training Setting} & \textbf{CLIP-I $\uparrow$} & \textbf{CSD $\uparrow$} & \textbf{AES $\uparrow$} & \textbf{Overall Mean $\uparrow$} \\
    \midrule
    Na\"ive E2E                                                     & 0.655             & \underline{0.455} & 0.668             & 0.593 \\
    w/o Subject pre-train                       & 0.726             & 0.433             & 0.678             & 0.612 \\
    w/o Style pre-train                          & \textbf{0.778}    & 0.399             & 0.676             & 0.618  \\
    Full Protocol w/o Masked Attention       & 0.758             & 0.452             & \underline{0.690} & \underline{0.633} \\
    \midrule
    \emph{Insert In Style} (Full Protocol + Masked Attention)       & \underline{0.761} & \textbf{0.466}    & \textbf{0.697}    & \textbf{0.641}  \\
    \bottomrule
    \end{tabular}
}
\label{tab:ablationstudy}
\end{table}

\begin{table}[t!]
\centering
\caption{Ablation on impact of our Prior Preservation Objective on Identity (CLIP-I) and Style (CSD)}
\resizebox{0.28\textwidth}{!}{
    \begin{tabular}{lcc}
    \toprule
    \textbf{Method} & \textbf{CLIP-I $\uparrow$} & \textbf{CSD $\uparrow$} \\
    \midrule
    \emph{Insert In Style}                     & \underline{0.761}     & \underline{0.466} \\
    w/o Identity prior preservation     & 0.731                 & 0.466 \\
    w/o Style prior Preservation        & \textbf{0.763}        & 0.432     \\
    \bottomrule
    \end{tabular}
}
\label{tab:ablationstudy_priorloss}
\end{table}

\section{Conclusion}
\label{sec:conclusion}
We introduced \emph{Insert In Style}, the first zero-shot generative framework for harmonious cross-domain object composition, solving the state-of-the-art's trade-off between practical ``blenders'' and impractical ``online generators''. Our novel multi-stage training protocol with masked-attention architecture and prior preservation loss are explicitly designed to manage competing identity and style signals, preventing the ``concept interference'' common in general-purpose models. Powered by our new $115$k sample dataset, the largest for this task, our method demonstrated state-of-the-art performance across all metrics and was strongly preferred by humans in a user study. We believe our framework, our human-calibrated data pipeline, and our new $788$ sample public benchmark, the largest for this task, open a new avenue for the under-explored task of cross-domain object insertion.

{
    \small
    \bibliographystyle{ieeenat_fullname}
    \bibliography{main}
}

\clearpage
\appendix

\twocolumn[
    \centering
    {\huge \textbf{Appendix}} \\
    \vspace{2em} 

    \addtocontents{toc}{\protect\setcounter{tocdepth}{2}} 
    \begin{minipage}{\textwidth}
    \hypersetup{linkcolor=black} 
    \tableofcontents
    \end{minipage}
    \vspace{2.0em}
]

\setcounter{figure}{0}
\setcounter{table}{0}
\renewcommand{\thefigure}{\thesection\arabic{figure}}
\renewcommand{\thetable}{\thesection\arabic{table}}

\clearpage
\section{Dataset Generation \& Filtering}

\begin{figure}[t!]
	\centering
	\includegraphics[width=0.99\linewidth]{images/supplementary_images/AcceptanceRate_vs_DataPurged.png}
    \caption{\textbf{Acceptance rate v/s purged data evaluated at each iteration during our dataset filtering process for dataset generated using LoRA-based and reference-based stylization methods. The process is terminated once acceptance rate is saturated, indicated by a star mark, beyond which further iterations do not improve the perceptual quality of the dataset.}}
    \label{fig:dataPurgedvsScore1}
\end{figure}

In this section, we extend our discussion on data generation and filtering. We employ two methods to stylise the base dataset $\mathcal{D}^{base}$ of approximately $40$k samples to obtain the cross-domain composition dataset of $\mathcal{D}^{sty}$ of approximately $350$k samples. Our first approach utilizes a LoRA-based stylization method, generating $200$k samples with $20$ distinct style archetypes, effectively introducing rich structural variations under popular stylistic transformations. Complementing this, the second method involves reference-based stylization, leveraging over $1,000$ distinct stylistic images from the Style30k dataset \cite{style30k_eccv2024} to create $150$k samples. Together, these methods ensure a comprehensive and balanced representation of both structural and textural variation.

We perform iterative human-in-the-loop filtering on $\mathcal{D}^{sty}$ to rigorously eliminate samples where the reference object exhibits significant identity loss or demonstrates incoherence with the overall image composition during the stylization process. At every iteration we identify rejection set $\mathcal{R}_t$, comprised of the union of the lowest deciles across all scoring dimensions. Human experts evaluate  ~$5\%$ samples from this set, providing their judgment on whether each sample should be accepted or rejected based on its quality. The iteration process continues until the samples consistently demonstrate high perceptual quality measured by high acceptance rate, at which point we terminate the filtering. We share the scores and the percentage of data purged for different style sources, including the stopping points in Fig.~\ref{fig:dataPurgedvsScore1}. This comprehensive approach guarantees that our final dataset $\mathcal{D}^{sty}_T$ of $115$k maintains high standards of structural integrity and visual consistency.

\section{\emph{Insert In Style Bench} Details}
\label{sec:insertinstylebenchdetails}
To establish a rigorous standard for cross-domain composition, we introduce \emph{Insert In Style Bench}. Comprising $788$ challenging pairs, it stands as the largest evaluation benchmark for this task, significantly surpassing TF-ICON \cite{tficon_iccv2023} ($332$), AIComposer \cite{aicomposer_iccv2025} ($367$), and SubjectPlop \cite{magicinsert_iccv2025} ($700$). Each pair includes a background and foreground image, accompanied by their respective masks. The benchmark integrates $51$ backgrounds meticulously curated from diverse public repositories, including Human-Art \cite{humanart_cvpr2023}, Wikiart \cite{wikiart_arxiv2015}, Kaggle \cite{kaggledb1,kaggledb2,kaggledb3,kaggledb4,kaggledb5}, and Pexels \cite{pexels}, with $25$ high-fidelity foreground concepts sourced from DreamBooth \cite{dreambooth_cvpr2023} and SOTA generative models \cite{gemini25_arxiv2025}. As illustrated in Fig.~\ref{fig:benchmark_background_images}, the background images span a vast artistic spectrum, including \emph{pencil sketch}, \emph{watercolor}, \emph{vector illustration}, \emph{oil painting}, \emph{neon}, and \emph{cartoon}, alongside complex styles that defy simple categorization. This remarkable diversity addresses a critical gap in prior benchmarks \cite{tficon_iccv2023, aicomposer_iccv2025, magicinsert_iccv2025}, which often lack stylistic variety.

As depicted in Fig.~\ref{fig:benchmark_foreground_images}, the foreground images encompass a broad spectrum of object classes, including `apple', `backpack', `berry bowl', `book', `burger', `camera', `can', `candle', `cat', `clock', `dog', `duck toy', `german shepherd dog', `grey slot plushie', `kite', `monster toy', `mug', `rat', `rc car', `robot toy', `straw hat', `teapot', `teddy bear', `vase', and `wolf plushie'. This comprehensive collection ensures robust evaluation across diverse scenarios, making \emph{Insert In Style Bench} a vital resource for advancing research in cross-domain image composition.

\begin{figure*}[t!]
	\centering
	\includegraphics[width=0.99\linewidth]{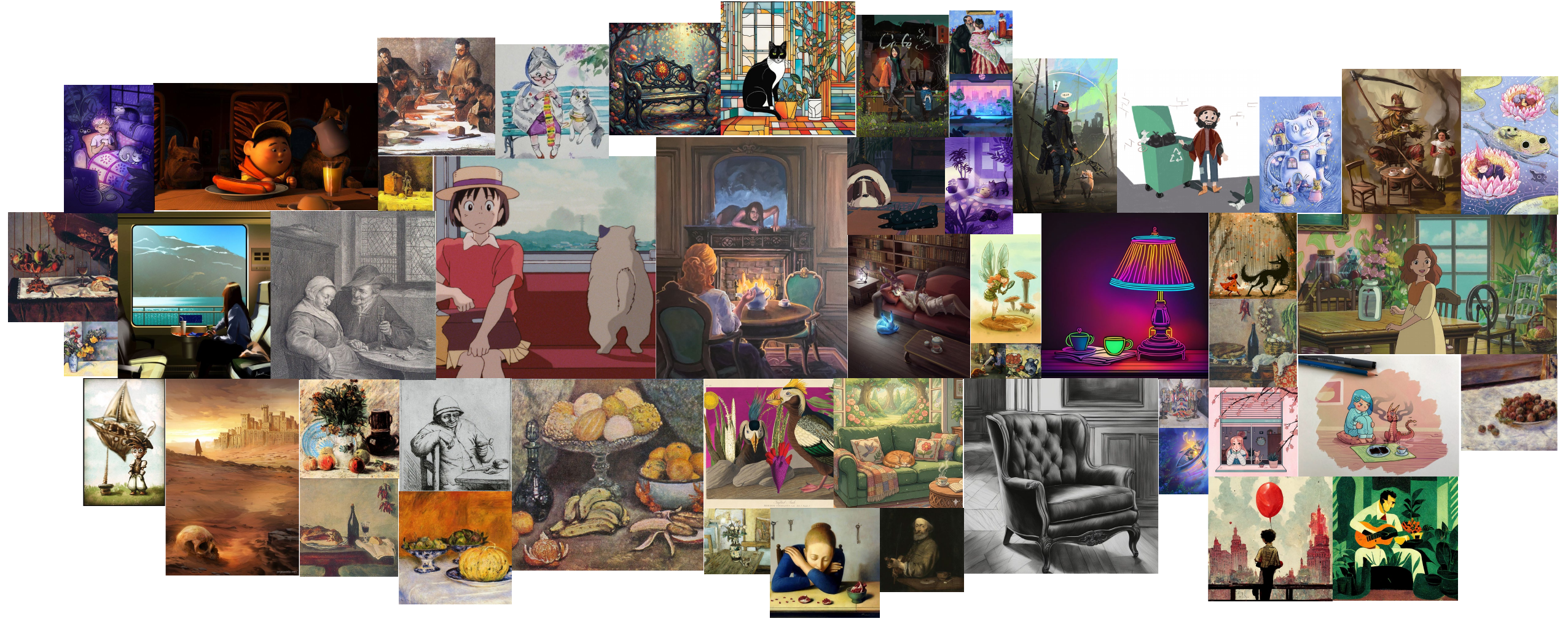}
	\caption{\textbf{Background diversity in \emph{Insert In Style Bench}.} We visualize the $51$ curated background scenes, spanning a vast artistic spectrum from vector illustrations and sketches to oil paintings and neon art. This extensive stylistic variety ensures a rigorous evaluation of cross-domain generalization.}
	\label{fig:benchmark_background_images}
\end{figure*}

\begin{figure*}[t!]
	\centering
	\includegraphics[width=0.99\linewidth]{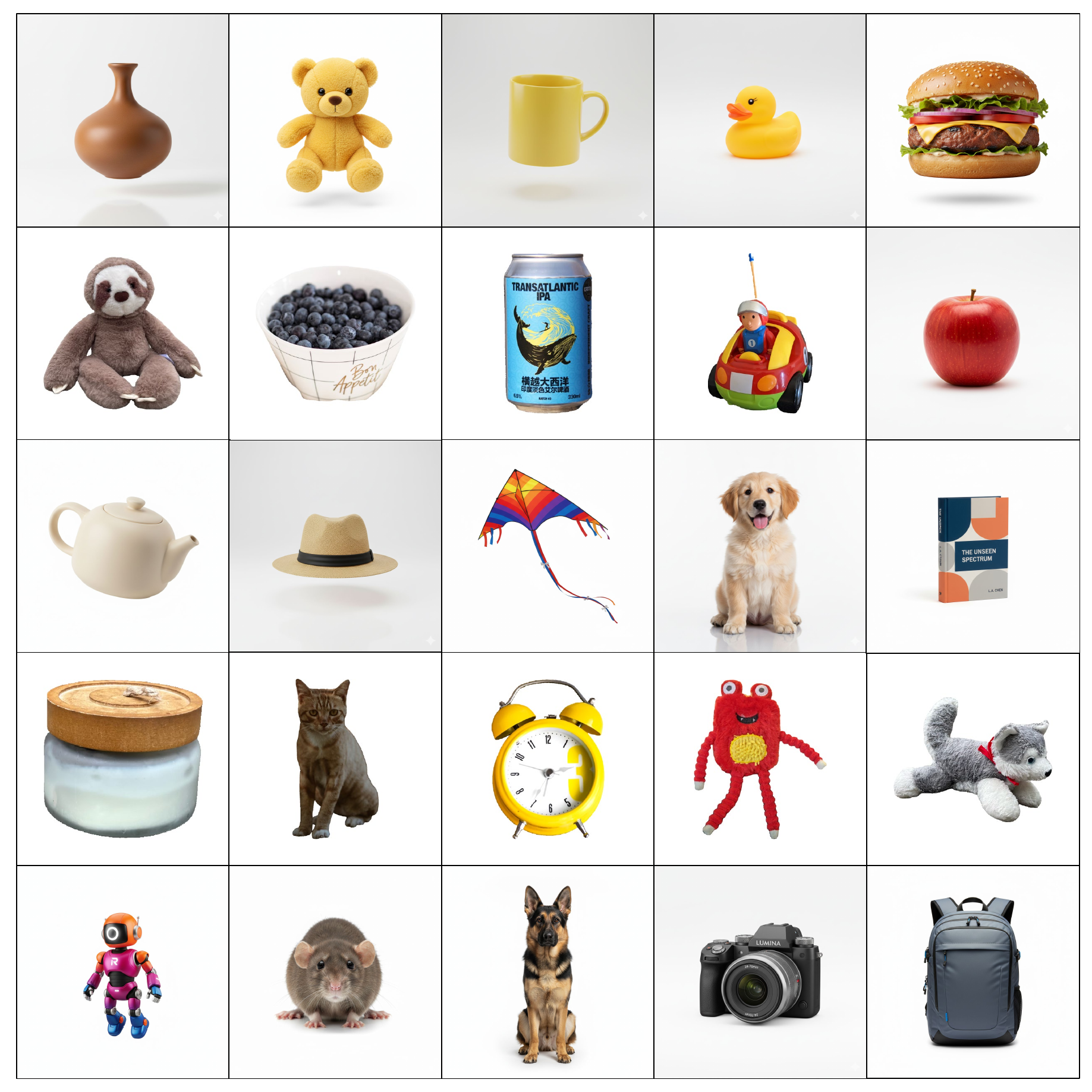}
	\caption{\textbf{Foreground concepts in \emph{Insert In Style Bench}.} We display the $25$ high-fidelity reference objects used for evaluation. The collection encompasses diverse semantic categories, including \emph{pets}, \emph{food}, \emph{toys}, and \emph{electronics}, to test identity preservation across varied structural complexities.}
	\label{fig:benchmark_foreground_images}
\end{figure*}

\section{Additional Ablation Studies}
\label{sec:additionalablationstudies}

\subsection{Ablation Study on Dataset Filtering}
\label{subsec:ablation_datasetfiltering}
To evaluate our filtering protocol, we trained models on datasets excluding either the identity consistency evaluation or style coherence evaluation during filtering. As shown in Fig.~\ref{fig:ablation_datasetfiltering}, the qualitative impact is distinct. Removing the identity-consistency evaluation degrades the model's ability to ground the subject, resulting in significant identity loss where the semantic features are not preserved. Conversely, removing the style coherence evaluation leads to foreground-background visual disconnection, where the object often appears ``pasted on'' with mismatched style and textures. Our full pipeline effectively removes these outliers during training, ensuring the final model learns both robust identity preservation and seamless style harmonization.

\begin{figure}[t!]
	\centering
	\includegraphics[width=0.99\linewidth]{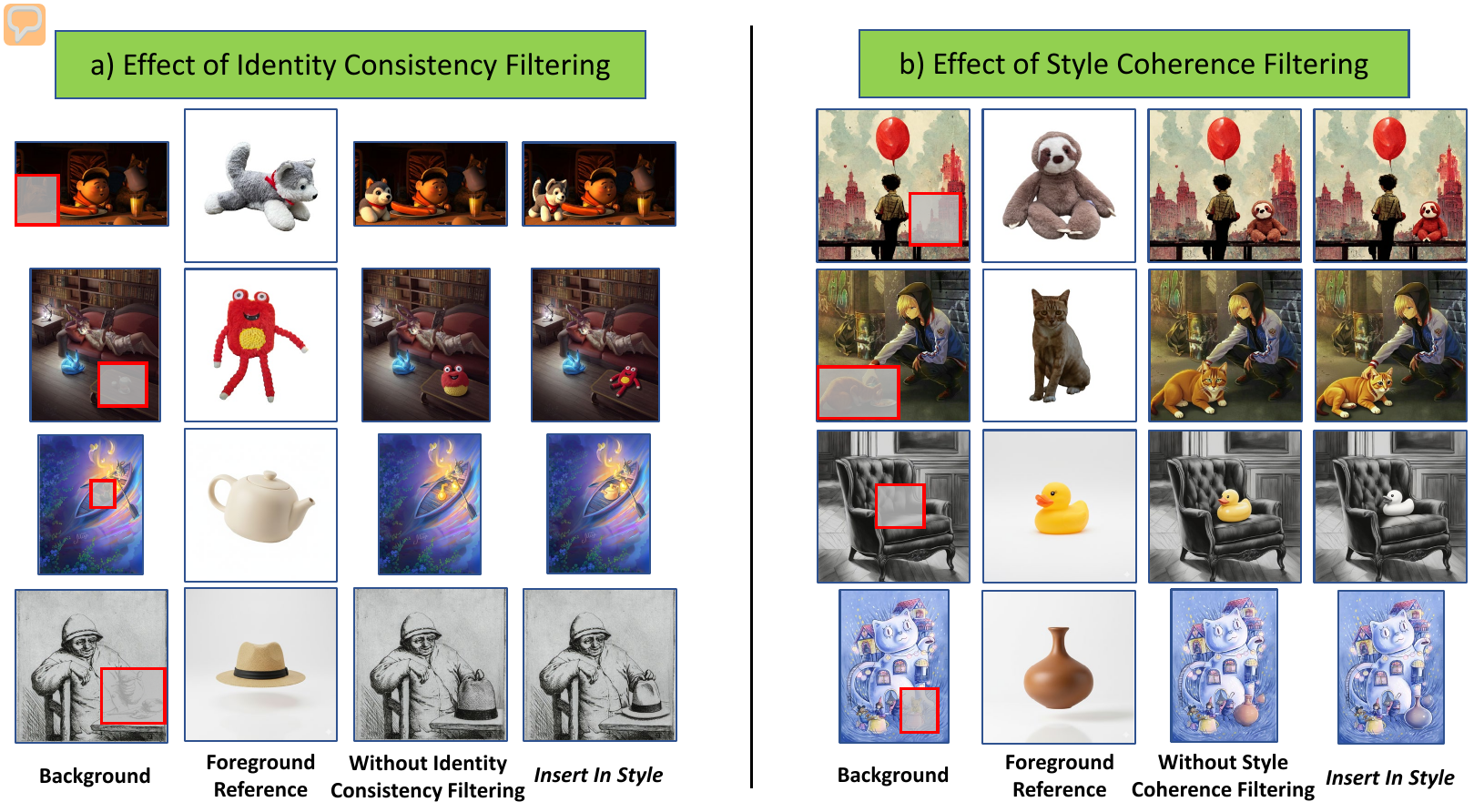}
	\caption{\textbf{Qualitative ablation of the dataset filtering pipeline.} \textbf{(a) Effect of Identity Evaluation:} Without the identity consistency evaluation while filtering, the model fails to preserve the object's semantic features, whereas our full pipeline maintains robust identity. \textbf{(b) Effect of Style Coherence Evaluation:} Without the style coherence evaluation while filtering, the object often appears ``pasted on'' with mismatched style and textures, whereas our full pipeline ensures the foreground is stylistically coherent with the background.}
	\label{fig:ablation_datasetfiltering}
\end{figure}

\subsection{Ablation Study on the Impact of Freezing Identity Branch and Style Branch}
\label{subsec:ablation_stage3unfreezing}

Our multi-stage training protocol relies on the premise that the \textbf{Identity} and \textbf{Style} branches act as specialized ``expert'' to learn robust disentangled representations. We hypothesize that freezing these branches in final compositional stage is essential to preserve their learned manifolds and prevent feature drift during the complex composition task.

To validate this, we compare our method against a baseline, where the pre-trained \textbf{Identity} and \textbf{Style} branches are left unfrozen during compositional stage training. Results in Tab.~\ref{tab:ablation_stage3unfreezing} and Fig.~\ref{fig:ablation_stage3unfreezedidentitystylebranches} confirm our design choice. The unfrozen baseline (Row $1$) suffers a significant degradation in both identity (CLIP-I) and style (CSD) metrics compared to our frozen protocol (Row $2$). This demonstrates that joint optimization leads to catastrophic forgetting of the specialized priors, whereas freezing along with prior preservation, as ablated in the main paper, ensures the model learns to compose the signals without corrupting the underlying representations.

\begin{table}[t!]
\centering
\caption{Ablation study on the impact of freezing identity and style branches in final composition stage training}
\resizebox{0.48\textwidth}{!}{
    \begin{tabular}{lcccc}
    \toprule
     \textbf{Training Strategy} & \textbf{CLIP-I $\uparrow$} & \textbf{CSD $\uparrow$} & \textbf{AES $\uparrow$} & \textbf{Overall Mean $\uparrow$}\\
    \midrule
    \makecell{Stage 3 w/ Unfreezed \\Identity and Style Branches}        &  0.752            & 0.464            & 0.695             & 0.637   \\
    \makecell{\emph{Insert In Style}}      & \textbf{0.761}    & \textbf{0.466}   & \textbf{0.697}    & \textbf{0.641}  \\
    \bottomrule
    \end{tabular}
}
\label{tab:ablation_stage3unfreezing}
\end{table}

\begin{figure}[t!]
	\centering
	\includegraphics[width=0.99\linewidth]{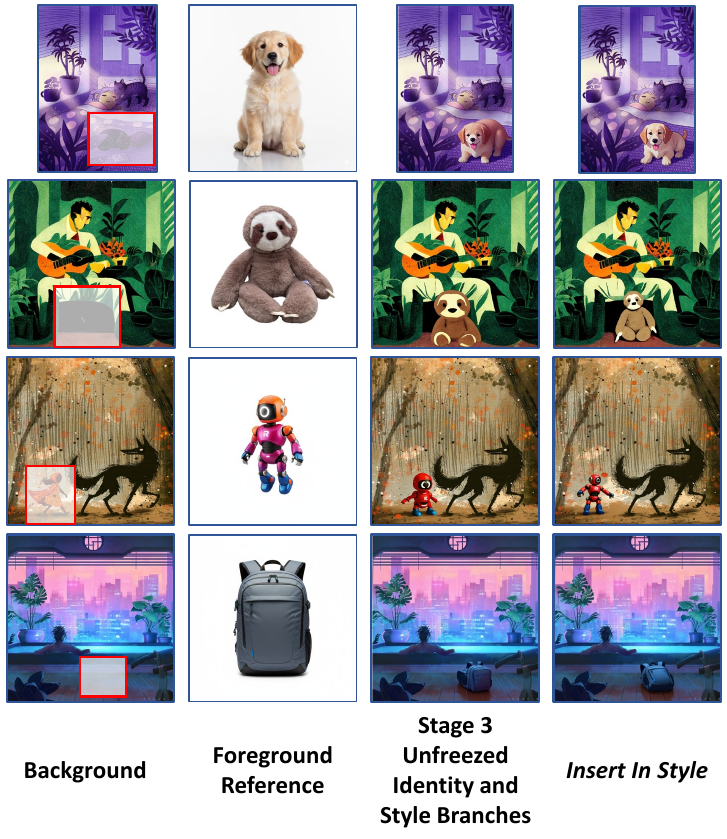}
	\caption{\textbf{Impact of freezing Identity and Style branches in Final Composite Model Training.} We compare our frozen protocol against a baseline, where the pre-trained \textbf{Identity} and \textbf{Style} branches are left unfrozen during final composite model training. The unfrozen baseline suffers from feature drift, degrading both object identity and style fidelity. In contrast, our frozen approach preserves the specialized priors, ensuring robust and harmonious composition.}
	\label{fig:ablation_stage3unfreezedidentitystylebranches}
\end{figure}

\section{Comparison against Instruction-based Editing (IBE) Methods}
\label{sec:ibe_evaluation}
We evaluated \emph{Insert In Style} against current state-of-the-art Instruction-based Editing (IBE) methods, including FLUX.$1$-Kontext, FLUX.$2$-Klein, and Qwen-Image-Edit. These evaluations were performed on a single A100 GPU with 80 GB of memory. To evaluate IBE methods, we extend \emph{Insert In Style} benchmark with human annotated instructions describing the location of the edit to accommodate the absence of background mask. Also, to indicate IBE methods to perform style harmonization of foreground reference with the background image we extend these human annotated instructions with ``while harmonizing and adapting to the background style and appropriate scale.''

As illustrated in Table~\ref{tab:ibe_benchmark_ours}, IBE methods demonstrate a strong capability in matching identity of the foreground objects, as shown by its high CLIP-I score. However, these methods fall short in maintaining style coherence with the background image (low CSD score). In contrast, \emph{Insert In Style} addresses this limitation by achieving superior results in both CLIP-I and CSD, thereby ensuring a more harmonious integration of the foreground object with the background.

Additionally, the qualitative results presented in Figure~\ref{fig:qual_ibe_iis} further highlight the capabilities of \emph{Insert In Style}. It is evident that existing IBE methods struggle with accurately positioning the foreground reference in the intended location and fail to maintain stylistic consistency with the background image. These shortcomings underscore the necessity of \emph{Insert In Style} for localized cross-domain insertion tasks, where precise placement and style coherence are critical.

\begin{table}[t!]
\centering
\caption{Quantitative comparison on the \emph{Insert In Style} benchmark using Instruction Based Editing methods. The best results are in \textbf{bold}, and the second best are \underline{underlined}.}
\resizebox{0.48\textwidth}{!}{
    \begin{tabular}{lccccc}
    \toprule
    \textbf{Method}                 & \textbf{CLIP-I $\uparrow$} & \textbf{CSD $\uparrow$} & \textbf{AES $\uparrow$} & \textbf{Overall Mean $\uparrow$} \\
    \midrule
    FLUX.1-Kontext                    & 0.665                   & \textbf{0.470}                      & 0.646                   & 0.594 \\
    FLUX.2-Klein 4B                   & \textbf{0.761}                   & 0.433                      & 0.679                   & 0.624 \\
    FLUX.2-Klein 9B                   & \underline{0.744}                   & 0.430                      & \underline{0.690}                   & 0.621 \\
    Qwen-Image-Edit                             & \underline{0.744}                   & 0.462                      & 0.689                   & \underline{0.632} \\
    \midrule
    \emph{Insert In Style}          & \textbf{0.761}                   & \underline{0.466}                      & \textbf{0.697}                   & \textbf{0.641} \\
    \bottomrule
    \end{tabular}
}
\label{tab:ibe_benchmark_ours}
\end{table}

\begin{figure*}[t!]
	\centering
	\includegraphics[width=0.99\linewidth]{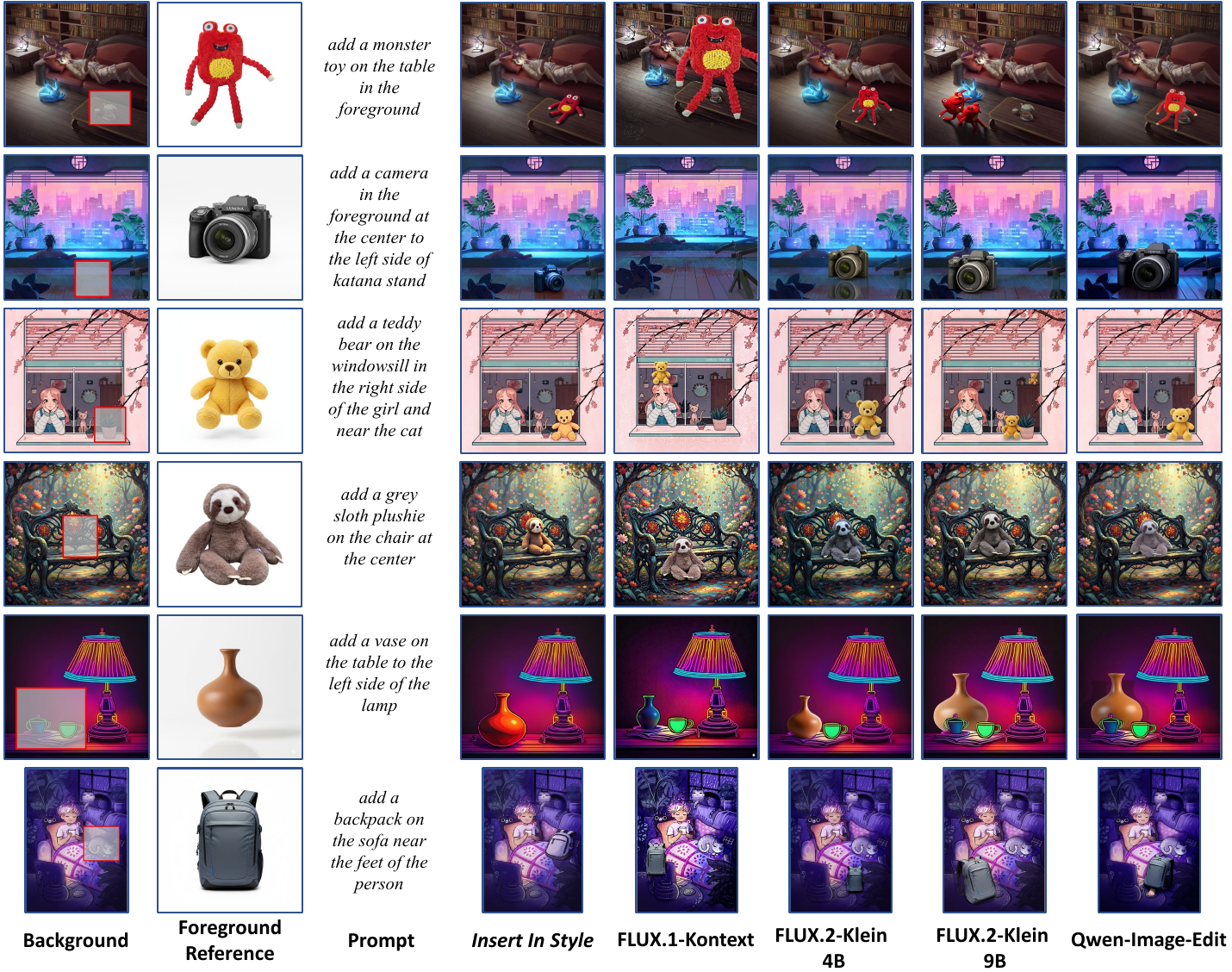}
	\caption{\textbf{Qualitative results on the \emph{Insert In Style} benchmark.} We compare our method against the Instruction-based Editing (IBE) methods like FLUX.1-Kontext, FLUX.2-Klein and Qwen-Image-Edit. We use the prompt with ``while harmonizing and adapting to the background style and appropriate scale'' to assist IBE methods in aligning the foreground reference with the background. Despite this, IBE methods still struggle with style coherence and precise spatial placement, whereas \emph{Insert In Style} consistently outperforms them.}
	\label{fig:qual_ibe_iis}
\end{figure*}

\section{Computational Cost Comparison}
\label{sec:computation_cost}
In {Tab.~\ref{tab:compute_cost} we compare the computational cost by measuring the average runtime, peak memory usage and number of image tokens used during inference of \emph{Insert In Style} against the baselines and IBE methods. All methods, including IBE models and baselines were evaluated using their default inference settings. {Tab.~\ref{tab:compute_cost} shows that \emph{Insert In Style} requires $24.59$ s per inference, which is $55\%$ of DiT based baseline DreamFuse's runtime ($44.2$ s). This is achieved while maintaining a comparable footprint of approximately $\sim33$ GB. \emph{Insert In Style} significantly outperforms Qwen-Image-Edit and Flux.2-Klein models in speed, a consequence of having fewer image tokens.

\begin{table}[h!]
\centering
\caption{Computational cost comparison of \emph{Insert In Style} against IBE-based methods and other baselines (`-' indicates non-DiT based methods with no image tokens).}
\resizebox{0.48\textwidth}{!}{
    \begin{tabular}{l|ccc}
    \toprule
    Method & Avg. Runtime (in secs) & Peak Memory Usage (in GB) & Image Tokens \\
    \midrule
    \textbf{Qwen-Image-Edit} & $161.84$ & $57.95$ & $12288$ \\
    \textbf{FLUX.2-Klein-9B} & $128.67$ & $34.75$ & $12288$ \\
    \textbf{FLUX.2-Klein-4B} & $64.37$ & $17.34$ & $12288$ \\
    \textbf{FLUX.1-dev}  & $6.08$  & $32.79$ & $4096$ \\
    \textbf{FLUX.1-Kontext-dev} & $28.47$ & $33.83$ & $8192$ \\ 
    \midrule
    \textbf{AnyDoor}  &  $8.05$  & $10.23$ & - \\
    \textbf{DreamFuse}  & $44.18$  & $34.58$ & $12288$ \\
    \textbf{TF-ICON} & $116.94$ & $18.38$ & - \\
    \textbf{TALE} & $6.92$ & $12.53$ & - \\
    \textbf{AIComposer} & $58.99$ & $19.81$ & - \\
    \midrule
    \textbf{\emph{Insert In Style}} & $24.59$ & $33.92$ & $8192$ \\
    \bottomrule
    \end{tabular}
}
\label{tab:compute_cost}
\end{table}

\section{Additional Benchmarking}
\label{sec:additionalbenchmark}

\begin{table}[t!]
\centering
\caption{Quantitative comparison on SubjectPlop dataset provided by Magic Insert \cite{magicinsert_iccv2025}. The best results are in \textbf{bold}, and the second best are \underline{underlined}.}
\resizebox{0.48\textwidth}{!}{
    \begin{tabular}{lccccccc}
    \toprule
    \textbf{Method}                 & \textbf{Venue} & \textbf{CLIP-I $\uparrow$} & \textbf{CSD $\uparrow$} & \textbf{AES $\uparrow$} & \textbf{Overall Mean $\uparrow$} \\
    \midrule
    AnyDoor                         & CVPR $2024$    & \textbf{0.834}      & 0.380               & 0.597              & 0.604 \\
    DreamFuse                       & ICCV $2025$    & \underline{0.823}   & 0.476               & \textbf{0.643 }    & \underline{0.647} \\ 
    \midrule
    TF-ICON                         & ICCV $2023$    & 0.755               & 0.385               & 0.588              & 0.576 \\
    TALE                            & ACMMM $2024$   & 0.734               & \textbf{0.499}      & 0.579              & 0.604 \\
    AIComposer                      & ICCV $2025$    & 0.814               & 0.476               & 0.611              & 0.634 \\
    \midrule
    \emph{Insert In Style}          & -              & 0.819               & \underline{0.484}   & \underline{0.638}  & \textbf{0.647} \\
    \bottomrule
    \end{tabular}
}
\label{tab:benchmark_subjectplop}
\end{table}

\begin{table}[t!]
\centering
\caption{Quantitative comparison on the in-domain object composition task using the AnyDoor \cite{anydoor_cvpr2024} benchmark dataset. The best results are in \textbf{bold}, and the second best are \underline{underlined}.}
\resizebox{0.48\textwidth}{!}{
    \begin{tabular}{lccccc}
    \toprule
    \textbf{Method}                 & \textbf{Venue} & \textbf{CLIP-I $\uparrow$} & \textbf{AES $\uparrow$} & \textbf{Overall Mean $\uparrow$} \\
    \midrule
    AnyDoor                         & CVPR $2024$    & \textbf{0.836}      & 0.473               & 0.654 \\
    DreamFuse                       & ICCV $2025$    & 0.754               & 0.505               & 0.629 \\ 
    \midrule
    TF-ICON                         & ICCV $2023$    & 0.738               & 0.480               & 0.609 \\
    TALE                            & ACMMM $2024$   & 0.762               & 0.476               & 0.619 \\
    AIComposer                      & ICCV $2025$    & 0.772               & \textbf{0.537}      & \underline{0.655} \\
    \midrule
    \emph{Insert In Style}          & -              & \underline{0.822}   & \underline{0.528}   & \textbf{0.675} \\
    \bottomrule
    \end{tabular}
}
\label{tab:benchmark_anydoor}
\end{table}

To further validate the robustness of \emph{Insert In Style}, we extend our evaluation to two additional benchmarks. We assess cross-domain fidelity using Magic Insert's SubjectPlop dataset \cite{magicinsert_iccv2025} and in-domain realism using the AnyDoor benchmark \cite{anydoor_cvpr2024}. For both domains, we present comprehensive quantitative and qualitative comparisons.

\subsection{SubjectPlop Benchmark}
\label{subsec:subjectplopbenchmark}
\paragraph{Note:} Due to the unavailability of the official Magic Insert \cite{magicinsert_iccv2025} code, we cannot report quantitative metrics for it. However, to ensure a rigorous comparison, we include qualitative results directly sourced from their official project website.

Quantitative results on the SubjectPlop benchmark  are presented in Tab.~\ref{tab:benchmark_subjectplop}. The trends mirror our main findings: in-domain specialists (DreamFuse \cite{dreamfuse_iccv2025}, AnyDoor \cite{anydoor_cvpr2024}) excel in identity (CLIP-I) but fail in stylization (CSD), while cross-domain baselines (TF-ICON \cite{tficon_iccv2023}, TALE \cite{tale_acmmm2024}) trade identity for style. However, a deeper qualitative analysis (Fig.~\ref{fig:qual_extra_subjectplop}) reveals critical nuances in these metrics.

We observe that AIComposer \cite{aicomposer_iccv2025} and TALE \cite{tale_acmmm2024} inflate their CSD scores via superficial color matching, i.e., blindly blending background hues into the foreground, rather than performing true structural stylization. This results in a ``copy-paste'' appearance where the object's texture remains photorealistic and incongruent. Similarly, DreamFuse \cite{dreamfuse_iccv2025} achieves a high AES score not by succeeding at the task, but by bypassing it: it generates aesthetically pleasing photorealistic compositions (Rows $1$, $2$, $5$) or hallucinates entirely new backgrounds (Row $4$), failing to adhere to the target style.

Fig.~\ref{fig:qual_extra_subjectplop} highlights a distinct qualitative gap between our method and the state-of-the-art generator Magic Insert \cite{magicinsert_iccv2025}. In challenging scenarios, Magic Insert \cite{magicinsert_iccv2025}, AIComposer \cite{aicomposer_iccv2025}, and TALE \cite{tale_acmmm2024} resort to aggressive color tinting—for example, turning a black-and-white animal reddish (Row $2$) or a character greenish (Row $1$) to match the background palette. This is a weak proxy for style transfer. In contrast, \emph{Insert In Style} performs structural stylization: it transforms the character’s rendering from $3$D to cartoon (Row $1$) and adapts the animal's fur texture (Row $2$) while correctly preserving intrinsic color and semantic identity.

Both qualitative and quantitative results showcase that \emph{Insert In Style} effectively bridges the gap between photorealism and artistic abstraction. Unlike baselines that struggle with the dichotomy of ``tinting vs. preserving", our method intelligently adapts surface texture and shading to the target domain while maintaining rigorous semantic fidelity, establishing a new standard for harmonious cross-domain composition.

\subsection{AnyDoor Benchmark}
\label{subsec:anydoorbenchmark}
To verify that our cross-domain capabilities do not compromise in-domain realism, we evaluate on the AnyDoor benchmark  in Tab.~\ref{tab:benchmark_anydoor} and Fig.~\ref{fig:qual_extra_anydoor}. We assess identity preservation (CLIP-I) and aesthetic quality (AES), applying our $20\%$ edit-mask filter to ensure metric validity.

Quantitative results in Tab.~\ref{tab:benchmark_anydoor} confirm that \emph{Insert In Style} remains highly competitive with specialized in-domain methods like DreamFuse \cite{dreamfuse_iccv2025} and AnyDoor \cite{anydoor_cvpr2024}. Qualitative comparisons in Fig.~\ref{fig:qual_extra_anydoor} corroborate this: our model produces photorealistic composites that match or even surpass baselines in terms of identity fidelity and composition (e.g., the `can' in Row $4$). Finally, while AIComposer \cite{aicomposer_iccv2025} achieves a high AES, our analysis reveals this is an artifact of aggressive over-harmonization: it achieves smooth blends by sacrificing the object's intrinsic color and identity, whereas our method preserves fidelity while ensuring realism.

\begin{figure*}[t!]
	\centering
	\includegraphics[width=0.99\linewidth]{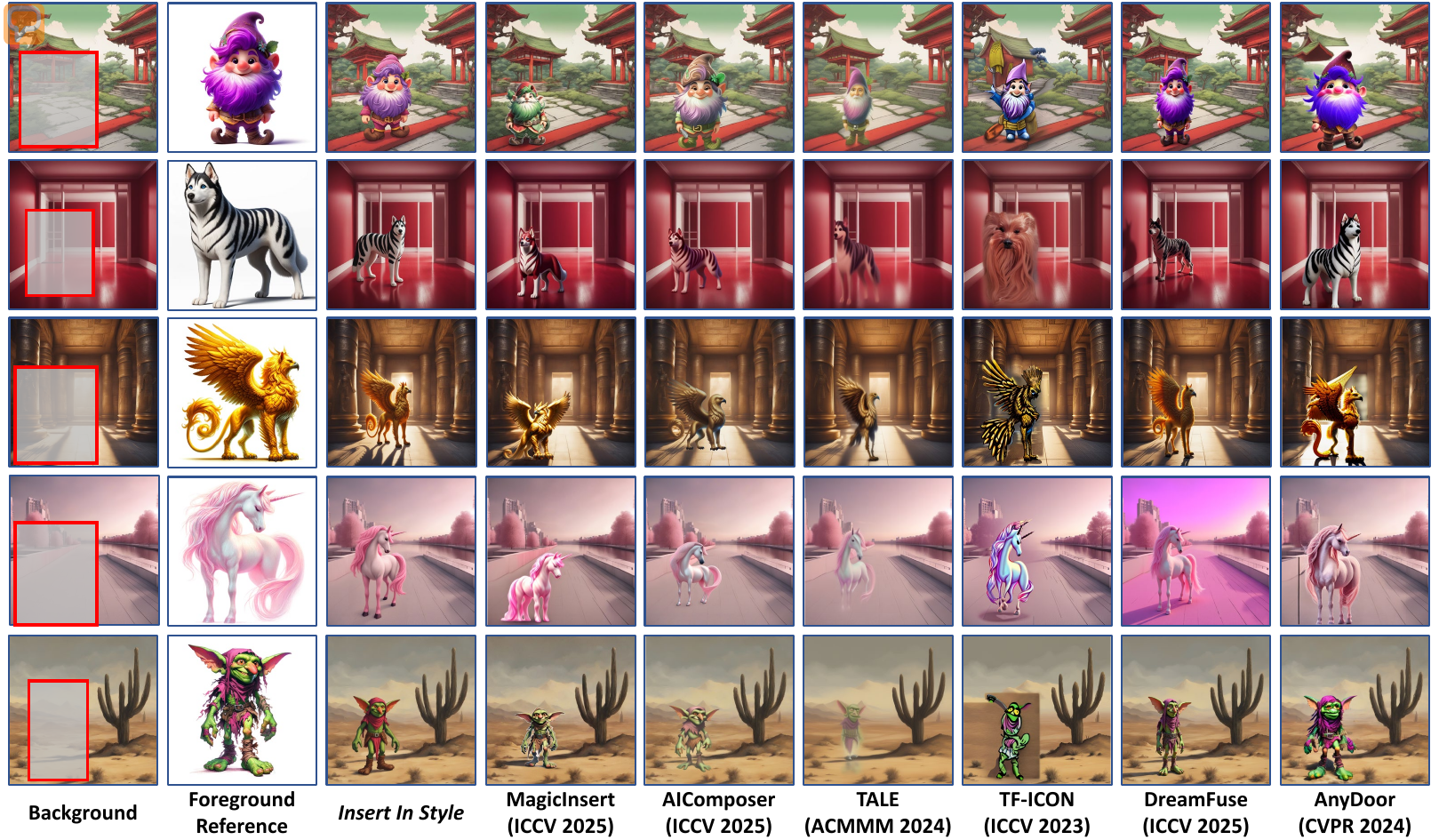}
	\caption{\textbf{Qualitative results on the SubjectPlop benchmark \cite{magicinsert_iccv2025}.} We compare against the state-of-the-art generator Magic Insert \cite{magicinsert_iccv2025} and blender AIComposer \cite{aicomposer_iccv2025}. While baselines rely on aggressive color tinting (e.g., turning the black-and-white animal reddish in Row $2$, and the character to greenish in Row $1$) to match the background, \emph{Insert In Style} performs true structural stylization, preserving the subject's semantic identity and intrinsic colors.}
	\label{fig:qual_extra_subjectplop}
\end{figure*}

\begin{figure*}[t!]
	\centering
	\includegraphics[width=0.99\linewidth]{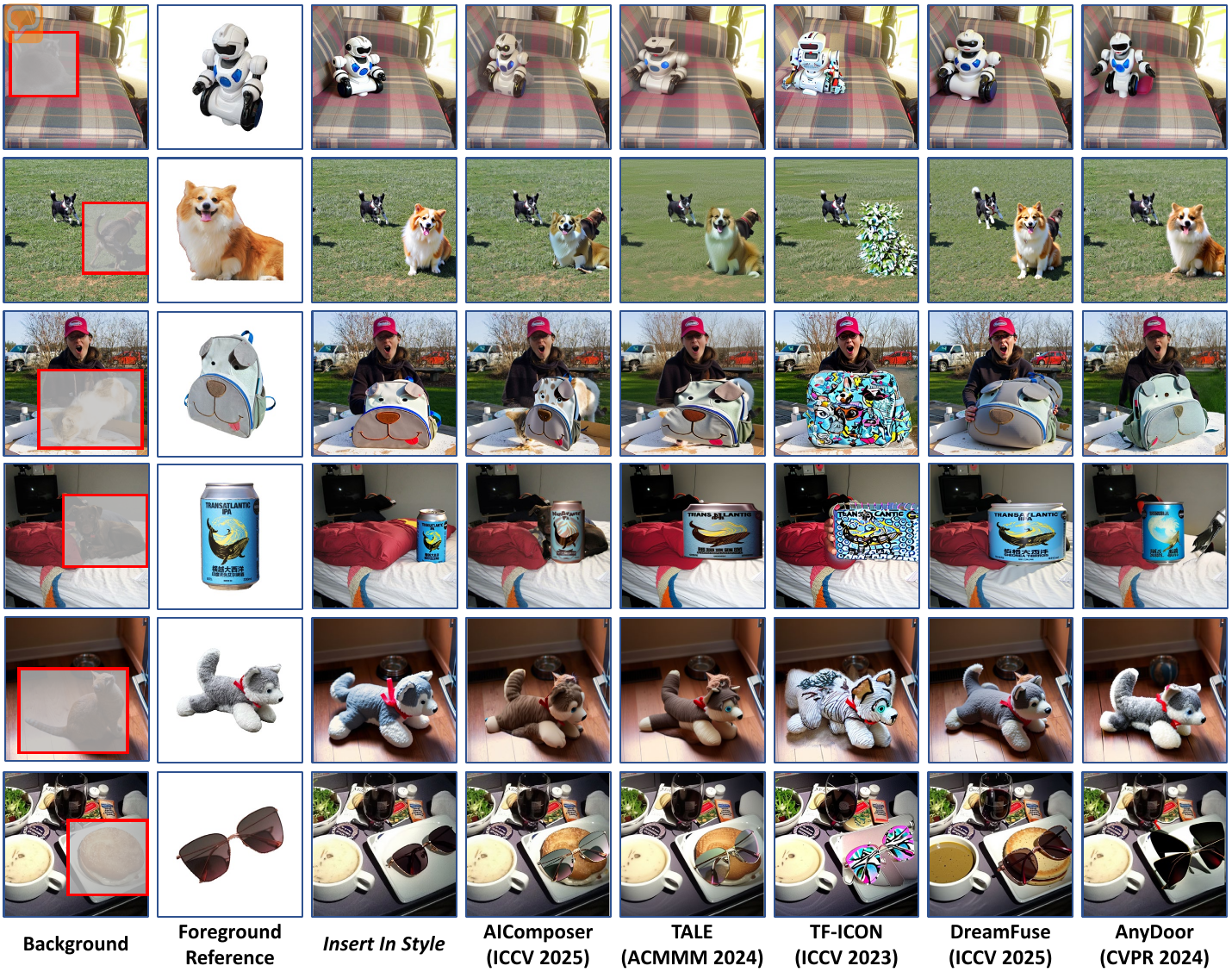}
	\caption{\textbf{Qualitative results on the in-domain AnyDoor benchmark \cite{anydoor_cvpr2024}.} \emph{Insert In Style} generates photorealistic composites competitive with the specialist AnyDoor \cite{anydoor_cvpr2024}, confirming that our cross-domain training does not compromise in-domain realism. In contrast, AIComposer \cite{aicomposer_iccv2025} sacrifices object fidelity for smooth blending, often losing texture details.}
	\label{fig:qual_extra_anydoor}
\end{figure*}

\section{Additional Qualitative Results}
\label{sec:additionalresults}
In this section, we provide an extensive gallery of qualitative results to further substantiate the robustness of \emph{Insert In Style}. We showcase comparisons across two diverse benchmarks: the AIComposer benchmark \cite{aicomposer_iccv2025} and our proposed \emph{Insert In Style Bench}. These additional examples reinforce our main findings regarding the limitations of existing methods and demonstrate our model's superior ability to balance identity fidelity with structural style adaptation.

\begin{figure*}[t!]
	\centering
	\includegraphics[width=0.99\linewidth]{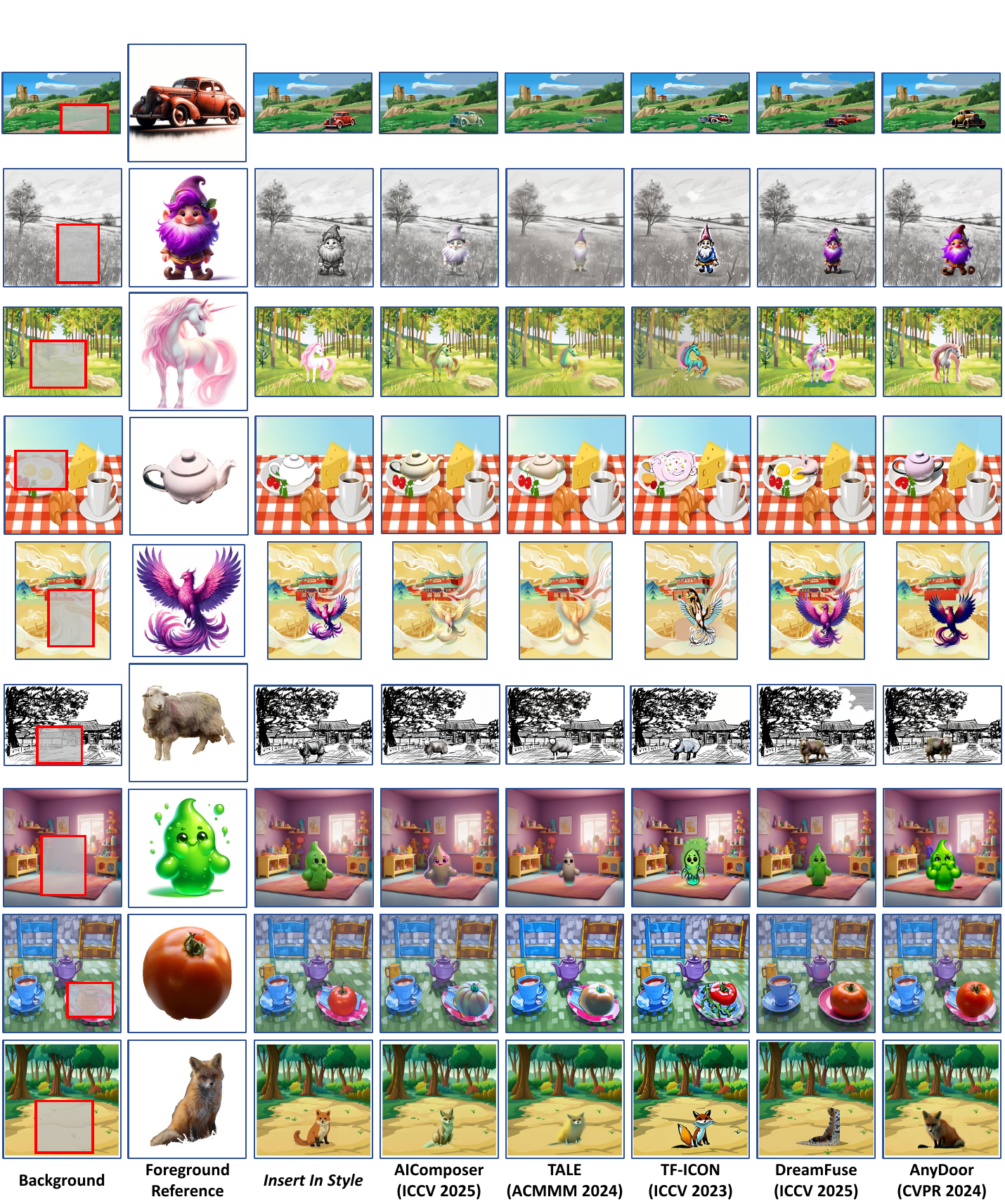}
	\caption{\textbf{Extended comparison on the AIComposer benchmark \cite{aicomposer_iccv2025}.} Consistent with our main analysis, AIComposer \cite{aicomposer_iccv2025} prioritizes global color alignment, resulting in ``pasted-on'' artifacts where the object's texture remains unadapted. \emph{Insert In Style} consistently achieves superior structural adaptation, ensuring the inserted subject is stylistically coherent with the scene.}
	\label{fig:qual_extra_aicomposer}
\end{figure*}

\begin{figure*}[t!]
	\centering
	\includegraphics[width=0.99\linewidth]{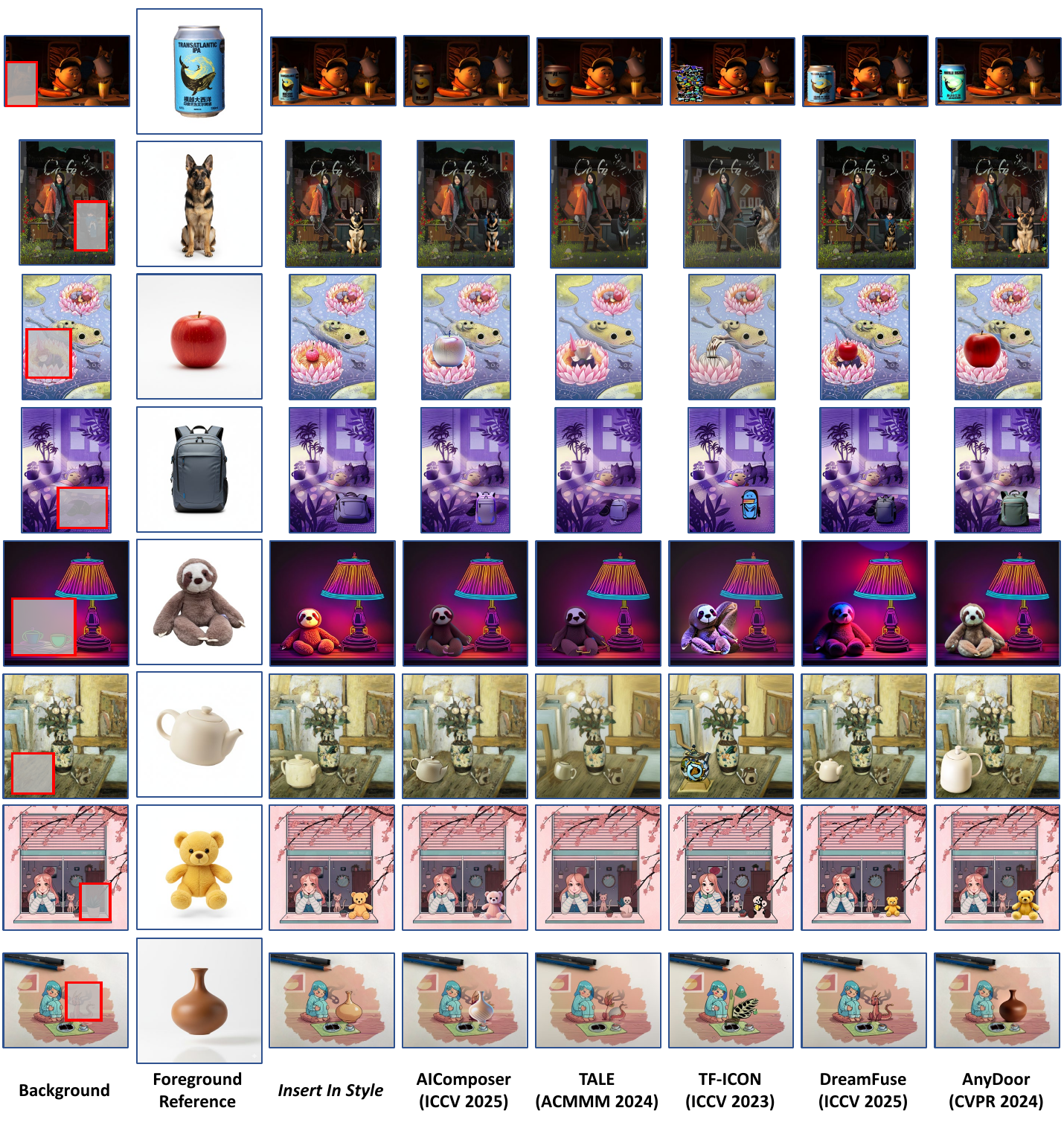}
	\caption{\textbf{Extended comparison on our \emph{Insert In Style Bench}.} Baselines exhibit color tampering (e.g., changing the teddy bear's color in Row $7$) to blindly match the background palette. \emph{Insert In Style} correctly distinguishes between style transfer and identity corruption, preserving intrinsic object semantics while applying the necessary artistic transformations.}
	\label{fig:qual_extra_ours}
\end{figure*}

\subsection{AIComposer Benchmark}
\label{subsec:additionalqualitative_aicomposer}
Fig.~\ref{fig:qual_extra_aicomposer} presents extended comparisons on the AIComposer \cite{aicomposer_iccv2025} benchmark. Consistent with our main analysis, baseline methods struggle to achieve true stylistic integration. Notably, AIComposer \cite{aicomposer_iccv2025} relies heavily on superficial color matching, i.e., tinting the object to match the background palette, while failing to adapt the object's texture or stroke style. This results in a jarring ``pasted-on'' appearance. In contrast, \emph{Insert In Style} consistently demonstrates superior structural adaptation, ensuring the inserted subject is both semantically faithful and stylistically coherent with the scene.

\subsection{Insert In Style Bench}
\label{subsec:additionalqualitative_insertinstylebench}
Fig.~\ref{fig:qual_extra_ours} showcases results on our \emph{Insert In Style Bench}. The comparisons in Rows $3$ (`apple'), $4$ (`backpack'), and $7$ (`teddy bear') are particularly revealing. In these interesting scenarios, cross-domain baselines like AIComposer \cite{aicomposer_iccv2025} exhibit color tampering, i.e., blindly overwriting the object's intrinsic colors with the background's dominant palette. This creates a fundamental loss of identity fidelity (e.g., an incorrect color shift for the `teddy bear'). Conversely, \emph{Insert In Style} preserves the object's intrinsic color semantics while successfully applying the necessary textural and artistic transformations, demonstrating a far more sophisticated understanding of cross-domain composition.

\section{Additional Capability: Customized Foreground Object Insertion using Text Prompt}
\label{sec:additionalqualitative_subjectcustomizationtextprompt}
A critical advantage of our generative framework over conventional ``blender'' methods is the ability to semantically modify the subject during insertion. While traditional approaches are limited to pixel-level composition, our zero-shot model leverages its generative prior to facilitate precise text-guided subject customization. We enable this at inference time by extending the default trigger word (\texttt{``this''}) to a composite instruction of the form: \texttt{``this <preposition> <customization>''}.

Fig.~\ref{fig:qual_additionalcapability} demonstrates the robustness of this capability. The model successfully synthesizes prompt-driven modifications while rigorously maintaining both object identity and global style coherence. 
\begin{itemize} 
    \item \textbf{Accessory Generation (Rows $1$-$2$):} We introduce external elements, such as \emph{sunglasses} and a \emph{scarf}. Crucially, these are not merely overlaid; they are generated with the correct lighting, texture, and artistic stroke style to perfectly match the background. 
    \item \textbf{Attribute Modification (Row $3$):} We modify intrinsic attributes (e.g., \emph{color}), effectively altering the subject's appearance without degrading its structural identity. 
    \item \textbf{Spatial Context Synthesis (Row $4$):} We demonstrate complex spatial reasoning with prompts like \texttt{``on a wooden stool''}. Here, the model generates entirely new geometry, i.e., a stool, that is geometrically valid and stylistically consistent with the scene, a generative feat impossible for copy-paste baselines. 
\end{itemize}

\begin{figure*}[t!]
	\centering
	\includegraphics[width=0.99\linewidth]{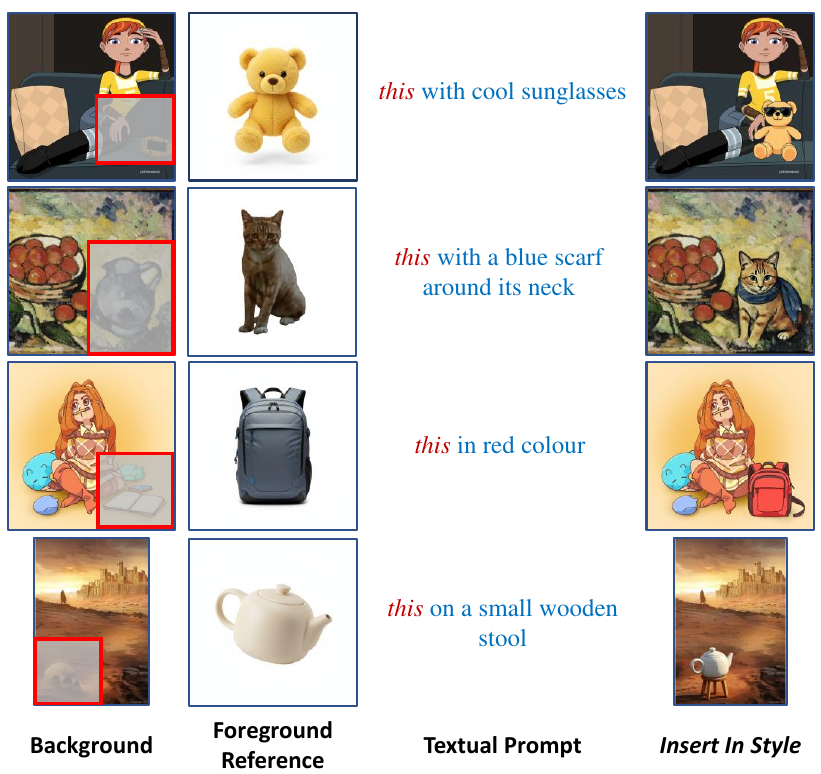}
	\caption{\textbf{Customized Foreground Object Insertion using Text Prompt.} Unlike conventional blenders, our generative framework enables precise semantic modifications via text prompts (e.g., \texttt{``this with sunglasses''}). In rows $1$-$2$, the model synthesizes accessories that perfectly match the scene's lighting and artistic style. In row $3$, it alters intrinsic attributes like color without degrading structural identity. In row $4$, it performs spatial context synthesis, generating entirely new, style-consistent geometry (a stool) to satisfy the positional constraint.}
	\label{fig:qual_additionalcapability}
\end{figure*}

\end{document}